\documentclass[lettersize,journal]{IEEEtran}
\usepackage{amsmath,amsfonts}
\usepackage{algorithmic}
\usepackage{algorithm}
\usepackage{array}
\usepackage[caption=false,font=normalsize,labelfont=sf,textfont=sf]{subfig}
\usepackage{textcomp}
\usepackage{stfloats}
\usepackage{url}
\usepackage{verbatim}
\usepackage{graphicx}
\usepackage{cite}
\usepackage{multirow}
\usepackage{color}
\usepackage{threeparttable}

\newcommand*{\red}{\textcolor{black}}

\begin{document}


\title{Aligning Cyber Space with Physical World: A Comprehensive Survey on Embodied AI}

\author{Yang Liu,~\IEEEmembership{Member,~IEEE}, Weixing Chen, Yongjie Bai, Xiaodan Liang,~\IEEEmembership{Senior~Member,~IEEE}, Guanbin Li, ~\IEEEmembership{Member,~IEEE}, Wen Gao,~\IEEEmembership{Fellow,~IEEE}, Liang Lin,~\IEEEmembership{Fellow,~IEEE}
\thanks{This work is supported in part by the National Key R\&D Program of China under Grant 2021ZD0111601; in part by the open research fund of Pengcheng Laboratory under Grant 2025KF1B0050; in part by the National Natural Science Foundation of China under Grant 62436009, Grant 62322608, and Grant 62301532; and in part by the Guangdong Basic and Applied Basic Research Foundation under Grant 2025A1515011874 and Grant 2023A1515011530. (\emph{Corresponding author: Liang Lin.})}
\thanks{Yang Liu, Weixing Chen, Yongjie Bai are with the School of Computer Science and Engineering, Sun Yat-sen University, China, and Guangdong Key Laboratory of Big Data Analysis and Processing, Guangzhou, China. (E-mail:liuy856@mail.sysu.edu.cn, chen867820261@gmail.com, baiyu8581@gmail.com)}
\thanks{Xiaodan Liang is with the Shenzhen Campus of Sun Yat-sen University, China,  Guangdong Key Laboratory of Big Data Analysis and Processing, Guangzhou, China, and Peng Cheng Laboratory, Shenzhen, China. (E-mail: xdliang328@gmail.com)}
\thanks{Guanbin Li and Liang Lin are with the School
of Computer Science and Engineering, Sun Yat-sen University, China, Guangdong Key Laboratory of Big Data Analysis and Processing, Guangzhou, China, and Peng Cheng Laboratory, Shenzhen, China. (E-mail: liguanbin@mail.sysu.edu.cn, linliang@ieee.org)}
\thanks{Wen Gao is with the Peng Cheng Laboratory, Shenzhen, China, and also with the Institute of Digital Media, Peking University, Beijing, China.  (E-mail:  wgao@pku.edu.cn)}
}
\markboth{IEEE/ASME Transactions on Mechatronics}%
{Shell \MakeLowercase{\textit{et al.}}: A Sample Article Using IEEEtran.cls for IEEE Journals}


\maketitle

\begin{abstract}
Embodied Artificial Intelligence (Embodied AI) is crucial for achieving Artificial General Intelligence (AGI) and serves as a foundation for various applications (e.g., intelligent mechatronics systems, smart manufacturing) that bridge cyberspace and the physical world. Recently, the emergence of Multi-modal Large Models (MLMs) and World Models (WMs) have attracted significant attention due to their remarkable perception, interaction, and reasoning capabilities, making them a promising architecture for embodied agents. In this survey, we give a comprehensive exploration of the latest advancements in Embodied AI. Our analysis firstly navigates through the forefront of representative works of embodied robots and simulators, to fully understand the research focuses and their limitations. Then, we analyze four main research targets: 1) embodied perception, 2) embodied interaction, 3) embodied agent, and 4) sim-to-real adaptation, covering state-of-the-art methods, essential paradigms, and comprehensive datasets. Additionally, we explore the complexities of MLMs in virtual and real embodied agents, highlighting their significance in facilitating interactions in digital and physical environments.  Finally, we summarize the challenges and limitations of embodied AI and discuss potential future directions. We hope this survey will serve as a foundational reference for the research community. The associated project can be found at \url{https://github.com/HCPLab-SYSU/Embodied_AI_Paper_List}.
\end{abstract}


\begin{figure}[t]
 \centering
\includegraphics[width=0.92\linewidth]{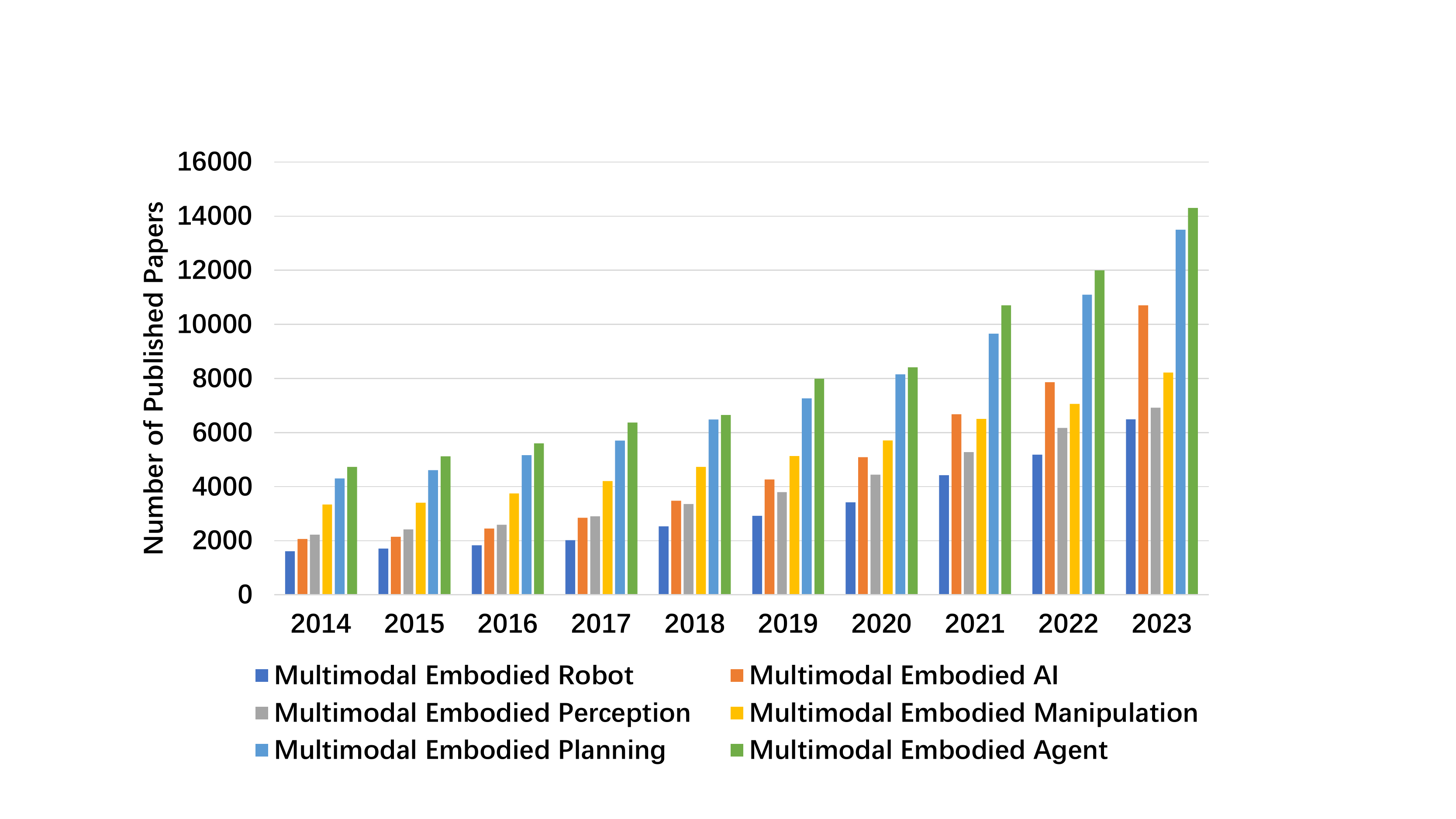}
   \vspace{-10pt}
    \caption{\red{The framework of the embodied agent based on MLMs and WMs, incorporates the ABC model, which stands for \textbf{A}I brain, \textbf{B}ody, and \textbf{C}ross-modal sensors.
    The embodied agent is equipped with an embodied world model as the A model, enabling it to understand the virtual-physical environment. Through the C model, it actively perceives multi-modal elements, enhancing its situational awareness. Meanwhile, the B model endows the agent execute actions, and interact with humans while utilizing tools effectively.
    }}
       \vspace{-10pt}
    \label{fig:framework}
\end{figure}

\section{Introduction}
\IEEEPARstart{E}{mbodied} AI was initially proposed from the Embodied Turing Test by Alan Turing in 1950 \cite{machinery1950computing} and has wide applications including robotics, healthcare, and smart manufacturing. Embodied AI is designed to determine whether agents can display intelligence that is not just limited to solving abstract problems in a virtual environment (cyber space\footnote{The agents are the foundation of both disembodied and embodied AI. The agents can exist in both cyber and physical spaces, integrated with various entities. The entities include not only robots but also other devices.}), but that is also capable of navigating the complexity and unpredictability of the physical world. For example, embodied AI enhances mechatronic systems by integrating with physical components for adaptive, real-world interactions, which enables systems to learn, perceive, and execute autonomously, improving efficiency and functionality, as shown in Fig. \ref{fig:framework}. The agents in the cyber space are generally referred to as disembodied AI, while those in the physical space are embodied AI (Table \ref{table:embodied}). Recent advances in Multi-modal Large Models (MLMs) have injected strong perception, interaction and planning capabilities to embodied models, to develop general-purpose embodied agents and robots that actively interact with virtual and physical environments. Therefore, the embodied agents are widely considered as the best carriers for MLMs. The recent representative embodied models are RT-2 \cite{zitkovich2023rt} and RT-H \cite{belkhale2024rt}. Nevertheless, the capabilities of long-term memory, understanding complex intentions, and the decomposition of complex tasks are limited for current MLMs.

To achieve Artificial General Intelligence (AGI), the development of embodied AI stands as a fundamental avenue.  Different from conversational agents like ChatGPT \cite{wu2023brief}, embodied AI believes that the true AGI can be achieved by controlling physical embodiments and interacting with both simulated and physical environments \cite{duan2022survey, ma2024survey}.  As we stand at the forefront of AGI-driven innovation, it is crucial to delve deeper into the realm of embodied AI, unraveling their complexities, evaluating their current developmental stage, and contemplating the potential trajectories they may follow in the future. Nowadays, embodied AI contains various key techniques across Computer Vision (CV), Natural Language Processing (NLP), and robotics, with the most representative being embodied perception, embodied interaction, embodied agents, and sim-to-real robotic control \cite{gao2023adaptive}. Therefore, it is imperative to capture the evolving landscape of embodied AI in the pursuit of AGI through a comprehensive survey.

\begin{table*}[t]
\caption{Comparison between disembodied AI and embodied AI.}
   \vspace{-8pt}
\centering
\begin{tabular}{c|c|c|c|c}
\hline
  Type & Environment& Physical Entities&Description&Representative Agents\\\hline
Disembodied AI&Cyber Space&No&Cognition and physical entities are disentangled&ChatGPT \cite{wu2023brief}, RoboGPT \cite{chen2023robogpt}\\
  Embodied AI&Physical Space&Robots, Cars, Other devices&Cognition is integrated into physical entities&RT-1 \cite{brohan2022rt}, RT-2 \cite{brohan2023rt}, RT-H \cite{belkhale2024rt}\\\hline
\end{tabular}
\vspace{-15pt}
\label{table:embodied}
\end{table*}

Embodied agent is the most prominent basis of embodied AI. For an embodied task, the embodied agent must fully understand the human intention in language instructions, actively explore the surrounding environments, comprehensively perceive the multi-modal elements from both virtual and physical environments, and execute appropriate actions for complex tasks \cite{hu2023toward,mccarthy2024towards}, as shown in Fig. \ref{fig:framework}. The rapid progress in multi-modal models exhibits superior versatility, dexterity, and generalizability in complex environments compared to traditional deep reinforcement learning approaches. Pre-trained visual representations from state-of-the-art vision encoders \cite{radford2021learning,li2023blip} provide precise estimations of object class, pose, and geometry, which makes the embodied models thoroughly perceive complex and dynamic environments. Powerful Large Language Models (LLMs) make robots better understand the linguistic instructions from humans. Promising MLMs give a feasible approach for aligning the visual and linguistic representations of embodied robots. The
world models \cite{assran2023self,zhu2024sora} exhibit remarkable simulation capabilities and promising comprehension of physical laws, which makes embodied models comprehensively understand both the physical and real environments.
These innovations empower embodied agents to comprehensively perceive complex environment, interact with humans naturally, and execute tasks reliably.

\red{Despite the intensive interest in harvesting the powerful perception and reasoning ability from MLMs, the research community is short of a comprehensive survey that can help sort out existing embodied AI studies, the challenges faced, as well as future research directions. In the era of MLMs, we aim to fill up this gap by performing a systematic survey of embodied AI across cyber space to physical world.
We conduct the survey from different perspectives including embodied robots, simulators, four representative embodied tasks (visual active perception, embodied interaction, multi-modal agents and sim-to-real adaptation), and future research directions. We believe that this survey will provide a clear big picture of what we have achieved, and we could further achieve along this emerging yet very prospective research direction.}

\red{\textbf{Differences from previous works}: Although there have been several survey papers \cite{pfeifer2004embodied,duan2022survey,ma2024survey,ren2024embodied} for embodied AI, most of them are outdated as they were published before the era of MLMs, which started around 2023. To the best of our knowledge, there are only two survey papers \cite{ma2024survey,ren2024embodied} after 2023, which focused on vision-language-action models and embodied AI system for smart manufacturing, respectively. Embodied AI, with AI brain, Body and Cross-modal sensors, is first proposed in \cite{ren2024embodied}, which is also the first work to propose technical architecture of embodied AI system for future smart manufacturing in the era of foundation model. However, the MLMs, WMs and embodied agents are not fully considered in previous surveys. Additionally, recent developments in embodied robots and simulators are also overlooked. To address the scarcity of comprehensive survey papers in this rapidly developing field, we propose this comprehensive survey that covers representative embodied robots, simulators, and four main research tasks: embodied perception, embodied interaction, embodied agents, and sim-to-real adaptation. In summary, the main contributions of this work are threefold:}

\begin{itemize}
    \item To the best of our knowledge, this is the first comprehensive survey of embodied AI from the perspective of the alignment of cyber and physical spaces based on MLMs and WMs, offering novel insights about methodologies, benchmarks, challenges, and applications.
    \item We categorize and summarize embodied AI into several essential parts including robots, simulators, and four main research tasks: embodied perception, embodied interaction, embodied agents and sim-to-real adaptation, which serve as a detailed taxonomy of embodied AI.
    \item To facilitate the development of robust, general-purpose embodied agents, we propose a new dataset standard ARIO (All Robots In One) and a unified large-scale ARIO dataset, encompassing approximately 3 million episodes collected from 258 series and 321,064 tasks.
\end{itemize}

\red{The rest of this survey is organized as follows. Section 2 introduces embodied robots. Section 3 describes general and real-scene embodied simulators. Section 4 introduces embodied perception, including active visual perception and visual language navigation. Section 5 introduces embodied interaction. Section 6 introduces embodied agents including the embodied multi-modal foundation model and embodied task planning. Section 7 introduces sim-to-real adaptation including embodied world model, data collection and training. In Section 8, we discuss promising research directions.}

\section{Embodied Robots}

\begin{figure}[t]
 \centering
\includegraphics[width=0.83\linewidth]{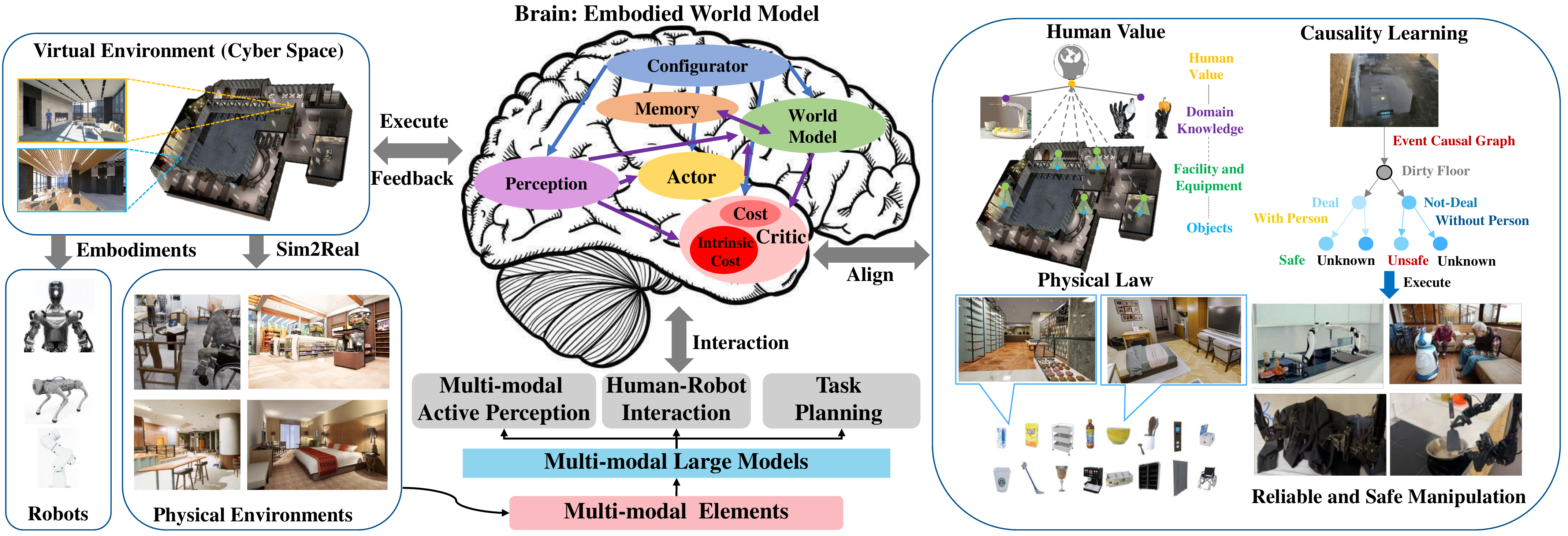}
   \vspace{-20pt}
    \caption{The Embodied Robots include Fixed-base Robots, Quadruped Robots, Humanoid Robots, Wheeled Robots, Tracked Robots, and Biomimetic Robots.}
           \vspace{-10pt}
    \label{fig:embodied_robot}
\end{figure}

Embodied agents interact with the physical environment, including robots, smart appliances, and autonomous vehicles, etc. Fixed-base robots, shown in Fig.~\ref{fig:embodied_robot} (a), are used in laboratory automation and industry due to their precision, e.g., Franka Emika panda \cite{liao2023dynamic,zhao2024dual}, Kuka iiwa \cite{li2024data,ortenzi2018vision}, and Sawyer \cite{kommuri2022external,spyrakos2020passivity}. Wheeled robots, depicted in Fig.~\ref{fig:embodied_robot} (b), are efficient in logistics and warehousing due to their simple structure and low cost, like Kiva and Jackal robots~\cite{wurman2008coordinating}. They face challenges on uneven terrain. Tracked robots, shown in Fig.~\ref{fig:embodied_robot} (c), are ideal for off-road tasks such as agriculture and disaster recovery. Their track systems provide stability on soft terrains~\cite{shao2024vision}
Quadruped robots, illustrated in Fig.~\ref{fig:embodied_robot} (d), excel in complex terrain exploration and rescue missions. Examples include Unitree Robotics' A1 and Go1, and Boston Dynamics Spot. Humanoid robots, shown in Fig.~\ref{fig:embodied_robot} (e), mimic human movements and behaviors to provide personalized services. Their dexterous hands enable them to perform intricate tasks~\cite{bennett2015multigrasp, chen2015mechanical}. With LLM, these robots are anticipated to enhance efficiency and safety in manufacturing, healthcare, and services~\cite{xiang2024language}. Biomimetic robots, depicted in Fig.~\ref{fig:embodied_robot} (f), replicate the movements and functions of natural organisms. This simulation aids in operating within complex environments and boosts energy efficiency by emulating biological mechanisms~\cite{siciliano2008springer,yang2022bioinspired}. Examples include fish-like~\cite{katzschmann2018exploration}, insect-like~\cite{de2022insect}, and soft-bodied robots~\cite{sinatra2019ultragentle}.

\section{Embodied Simulators}



Embodied simulators are crucial for embodied AI due to their cost-effectiveness, safety features, scalability, rapid prototyping, and accessibility for research. They allow for controlled experimentation, data generation for training and evaluation, and standardized benchmarks. To facilitate interaction with the environment, it's essential to build realistic simulations by considering physical characteristics, object properties, and their interactions. This section will introduce the commonly used simulation platforms in two parts: the general simulator based on underlying simulation and the simulator based on real scenes.

\begin{table*}\scriptsize
\caption{General Simulator. \textbf{HFPS}: high-fidelity physical simulation; \textbf{HQGR}: high-quality graphics rendering; \textbf{RRL}: rich robot library; \textbf{DLS}: deep learning support; \textbf{LSPC}: large-scale parallel computing; \textbf{ROS}: tight integration with ROS; \textbf{MSS}: multiple sensor simulation; \textbf{CP}: cross-platform \textbf{Nav}: robot navigation \textbf{AD}: auto driving; \textbf{RL}: reinforcement learning  \textbf{LSPS}: large-scale parallel sim \textbf{MR}: multi-robot systems \textbf{RS}: robot simulation. $\circ$ indicates that the simulator excels at this aspect. }
       \vspace{-15pt}
\centering\setlength{\tabcolsep}{0.8mm}
    \begin{threeparttable}
    \red{\begin{tabular}{p{2.7cm}lp{1cm}p{1cm}p{1cm}p{1cm}p{1cm}p{1cm}p{1cm}p{1cm}p{3cm}p{2cm}}
    \hline
         Simulator& Year&\textbf{HFPS}& \textbf{HQGR}	& \textbf{RRL}	& \textbf{DLS}	& \textbf{LSPC} & \textbf{ROS}	& \textbf{MSS} & \textbf{CP} &  Physics Engine& Main Applications\\
             \hline
         Genesis \cite{Genesis} &  2024&$\circ$&$\circ$ &$\circ$ & $\circ$& $\circ$& $ $& $\circ$& $\circ$&  Custom& RL, LSPS, RS\\
         Isaac Sim \cite{isaacsim} &  2023&$\circ$&$\circ$ &$\circ$ & $\circ$& $\circ$& $\circ$& $\circ$& $\circ$&  PhysX& Nav, AD\\
         Isaac Gym \cite{isaacgym}&  2019&& $\circ$& & $\circ$& $\circ$& & & &  PhysX& RL,LSPS\\
         Gazebo \cite{gazebo}&  2004&& $\circ$&$\circ$ & & & $\circ$& $\circ$& $\circ$&  ODE, Bullet, Simbody, DART& Nav,MR\\
         PyBullet \cite{coumans2016pybullet}&  2017&& & & $\circ$& & & & $\circ$& Bullet&RL,RS\\
         Webots \cite{webots} &  1996&& $\circ$&$\circ$& & & & $\circ$& $\circ$& ODE&RS\\
         MuJoCo \cite{todorov2012mujoco}&  2012&$\circ$& & & $\circ$& & & & $\circ$& Custom &RL, RS\\
         Unity ML-Agents \cite{juliani}&  2017&&$\circ$ & & $\circ$& & & & $\circ$& Custom &RL, RS\\
         AirSim \cite{airsim2017fsr}&  2017&& $\circ$& & & & & & $\circ$& Custom &Drone sim, AD, RL\\
         MORSE \cite{morse}&  2015&& & & & & & $\circ$& $\circ$& Bullet&Nav, MR\\
         V-REP (CoppeliaSim) \cite{rohmer2013v}&  2013&& $\circ$&$\circ$ & & & & $\circ$& $\circ$& Bullet, ODE, Vortex, Newton&MR, RS\\    \hline
    \end{tabular}}
    \vspace{-15pt}
    \label{table:General Simulator}
    \end{threeparttable}
\end{table*}

\subsection{General Simulator}
The physical interactions and dynamic changes present in real environments are irreplaceable. However, deploying embodied models in the physical world often incurs high costs and faces numerous challenges. General-purpose simulators provide a virtual environment that closely mimics the physical world, allowing for algorithm development and model training, which offers significant cost, time, and safety advantages.

\begin{figure}[!t]
    \centering
    \includegraphics[width=1\linewidth]{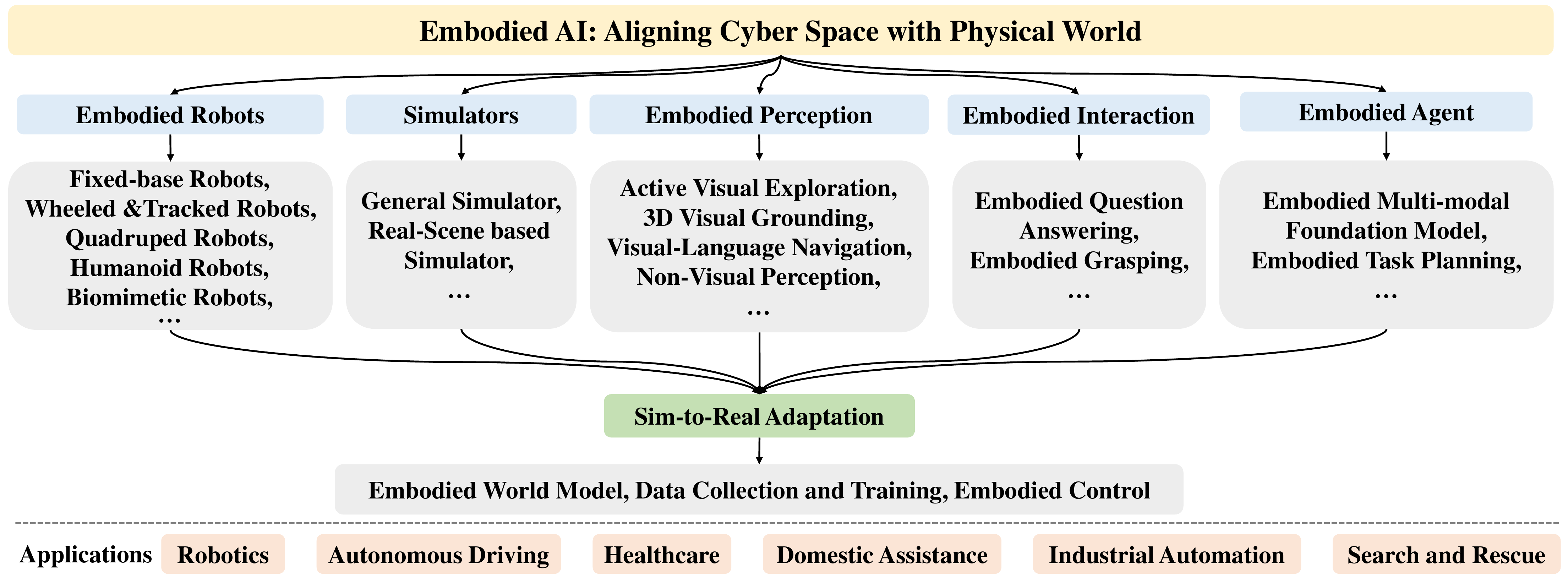}
       \vspace{-20pt}
    \caption{\red{Examples of General Simulators. The MuJoCo's figure is from \cite{wang2020learning}.}}
       \vspace{-15pt}
    \label{fig:General Simulator}
\end{figure}

\textbf{Isaac Sim} \cite{isaacsim} is an advanced simulation platform for robotics and AI research. It has high-fidelity physical simulation, real-time ray tracing, an extensive library of robotic models, and deep learning support. Its application scenarios include autonomous driving, industrial automation, and human-robot interaction. \textbf{Gazebo} \cite{koenig2004design} is an open-source simulator for robotics research. It has extensive robot libraries, and tight integration with ROS. It supports the simulation of various sensors and offers numerous pre-built robot models and environments. It is mainly used for robot navigation and control and multi-robot systems. \textbf{PyBullet} \cite{coumans2016pybullet} is the python interface for the Bullet physics engine. It is easy to be used and has diverse sensor simulation and deep learning integration. PyBullet supports real-time physical simulation, including rigid body dynamics, collision detection, and constraint solving. \red{Moreover, the newly launched \textbf{Genesis} \cite{Genesis} has differentiable physics engine and impressive generative capabilities. Table. \ref{table:General Simulator} presents the key features and primary application scenarios of 11 general-purpose simulators. Fig. \ref{fig:General Simulator} shows the visualization effects of the general simulators. }

\subsection{Real-Scene Based Simulators}

Achieving universal embodied agents in household activities is a primary focus in embodied AI. These embodied agents need to deeply understand human daily life and perform complex embodied tasks such as navigation and interaction in indoor environments. To meet the demands of these complex tasks, the simulated environments need to be close to the real world, which places high demands on the complexity and realism of the simulators. These simulators mostly collect data from the real world, create photorealistic 3D assets, and build scenes using 3D game engines like UE5 and Unity. The rich and realistic scenes make simulators based on real world environments the top choice in household activities. 
\begin{figure}[!t]
    \centering
    \includegraphics[width=0.95\linewidth]{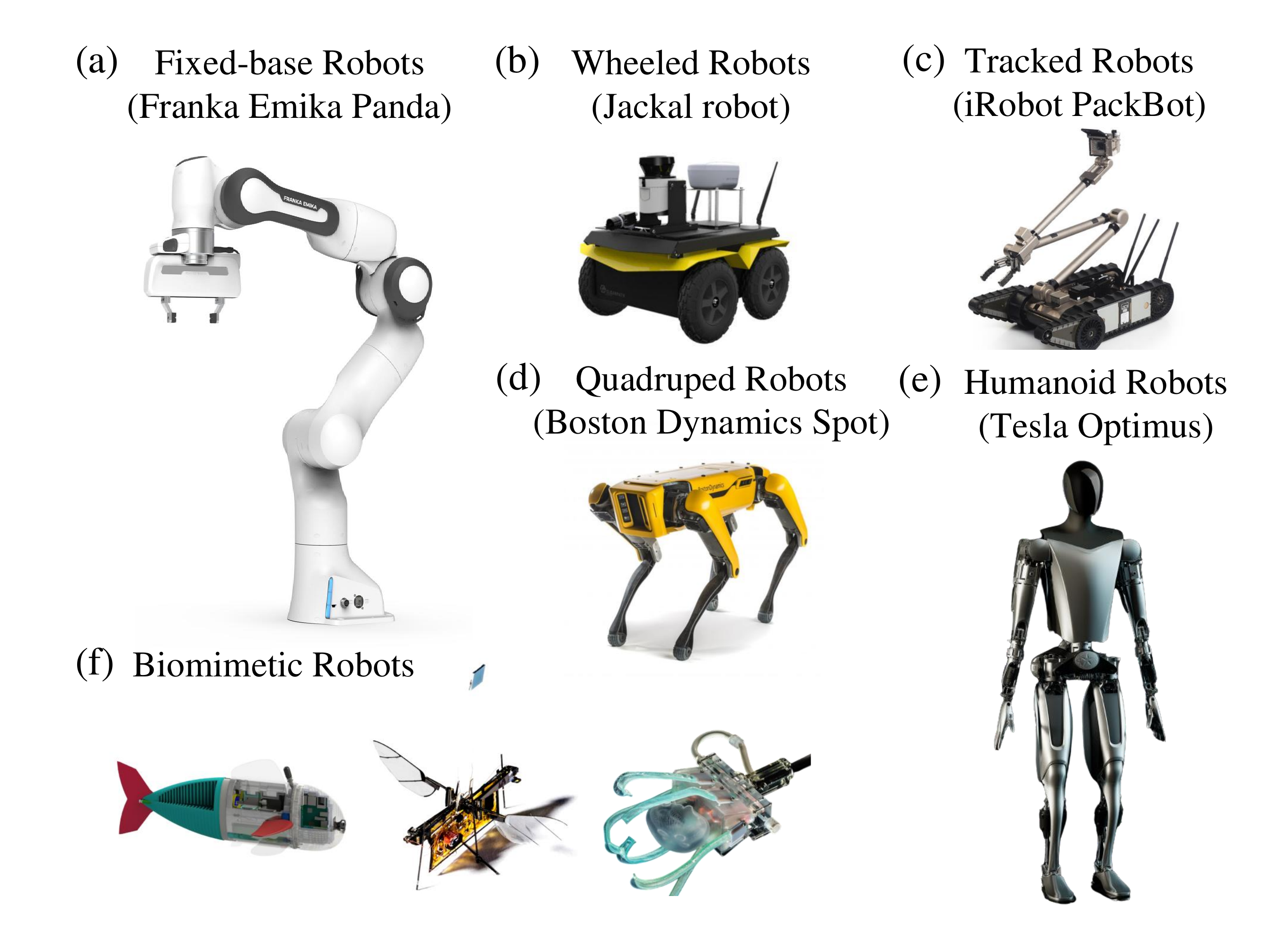}
       \vspace{-10pt}
    \caption{\red{Examples of Real-Scene Based Simulators.}}
       \vspace{-16pt}
    \label{fig:Real-Scene Based Simulators}
\end{figure}

\textbf{SAPIEN} \cite{SAPIEN} stands out for its design, specifically tailored to simulate interactions with joint objects like doors, cabinets, and drawers. \textbf{VirtualHome} \cite{virtualhome} is notable for its unique environment graph, which facilitates high-level embodied planning based on natural language descriptions of environments. While \textbf{AI2-ThOR} \cite{AI2-THOR} offers a wealth of interactive scenes, these interactions, similar to those in VirtualHome, are script-based and lack real physical interactions. This design suffices for embodied tasks not requiring fine-grained interactions. Both \textbf{iGibson} \cite{iGibson1} and \textbf{TDW} \cite{TDW} provide fine-grained embodied control and highly simulated physical interactions. iGibson excels in offering abundant and realistic large-scale scenes, making it suitable for complex and long-term mobile operations, whereas TDW allows greater user freedom in scene expansion and features unique audio and flexible fluid simulations, making it indispensable for related simulation scenarios. \textbf{Matterport3D} \cite{Matterport3D}, a foundational 2D-3D visual dataset, is widely used and extended in embodied AI benchmarks. Although the embodied agent in Habitat lacks interaction capabilities, its extensive indoor scenes, user-friendly interfaces, and open framework make it highly regarded in embodied navigation. \red{\textbf{InfiniteWorld} \cite{ren2024infiniteworld} focuses on unified and scalable simulation framework and implemented various improvements and the latest implicit asset reconstruction, as well as natural language-driven scene generation and editing. It provides strong support for complex robotic interactions through distributed collaboration, AI assistance, and Human-in-the-Loop.}

Besides, automated simulation scene construction is greatly beneficial for obtaining high-quality embodied data. \textbf{RoboGen} \cite{RoboGen} customizes tasks from randomly sampled 3D assets through LLMs, thereby creating scenes and automatically training agents; \textbf{HOLODECK} \cite{yang2024holodeck} can automatically customize corresponding high-quality simulation scenes in AI2-THOR based on human instructions; \textbf{PhyScene} \cite{yang2024physcene} generates interactive and physically consistent high-quality 3D scenes based on conditional diffusion. The Allen Institute for Artificial Intelligence expanded AI2-THOR and proposed \textbf{ProcTHOR} \cite{procthor}, which can automatically generate simulated scenes with sufficient interactivity, diversity, and rationality.

\section{Embodied Perception} 


The ``north stars" of the future of visual perception is embodied-centric visual reasoning and social intelligence \cite{fei2022searching}. Unlike merely recognizing objects in images, agent with embodied perception must move in the physical world and interact with the environment. This requires a deeper understanding of 3D space and dynamic environments. Embodied perception requires visual perception and reasoning, understanding the 3D relations within a scene, and predicting and performing complex tasks based on visual information.

\subsection{Active Visual Perception}

\begin{table*}[t]
\caption{The comparison of the active visual perception methods.}
\vspace{-8pt}
\scriptsize
\centering
\begin{tabular}{c|c|c}
\hline
Function & Type & Methods \\ \hline
\multirow{4}{*}{vSLAM} & \multirow{1}{*}{Traditional vSLAM} & MonoSLAM\cite{davison2007monoslam}, ORB-SLAM\cite{mur2015orb}, LSD-SLAM\cite{engel2014lsd}    \\  \cline{2-3}

 & \multirow{2}{*}{Semantic vSLAM} &  SLAM++\cite{salas2013slam++}, 
 QuadricSLAM\cite{nicholson2018quadricslam}, So-SLAM\cite{liao2022so},  \\
 &  & 
 SG-SLAM\cite{cheng2022sg}, OVD-SLAM\cite{he2023ovd}, GS-SLAM\cite{yan2024gs}     \\ \hline

\multirow{4}{*}{3D Scene Understanding} & \multirow{1}{*}{Projection-based} &  MV3D\cite{chen2017multi}, PointPillars\cite{lang2019pointpillars}, MVCNN\cite{su2015multi}       \\ \cline{2-3}
 & \multirow{1}{*}{Voxel-based} &  VoxNet\cite{maturana2015voxnet}, SSCNet\cite{song2017semantic}), MinkowskiNet\cite{choy20194d}, SSCNs\cite{graham20183d}, Embodiedscan\cite{wang2024embodiedscan}   \\ \cline{2-3}
 & \multirow{2}{*}{Point-based} &  PointNet\cite{qi2017pointnet}, PointNet++\cite{qi2017pointnet++}, PointMLP\cite{ma2022rethinking}),  PointTransformer\cite{zhao2021point}, Swin3d\cite{yang2023swin3d}, PT2\cite{wu2022point},
 \\
 &  &  
 3D-VisTA\cite{zhu20233d}, LEO\cite{huang2023embodied},
 PQ3D\cite{zhu2024unifying},
 PointMamba\cite{liang2024pointmamba}, 
 Mamba3D\cite{han2024mamba3d}   \\ \hline
\multirow{2}{*}{Active Exploration} & \multirow{1}{*}{Interacting with the environment} &   Pinto et al.\cite{pinto2016curious}, Tatiya et al.\cite{tatiya2023transferring}       \\ \cline{2-3}
 & \multirow{1}{*}{Changing the viewing direction} & Jayaraman et al. \cite{jayaraman2018learning}, NeU-NBV\cite{jin2023neu}, Hu et al.\cite{hu2023off}, Fan et al.\cite{fan2024evidential}    \\ \hline
\end{tabular}
\vspace{-15pt}
\label{tab:Active-Visual-Perception-methods}
\end{table*}

\begin{figure}
    \centering
    \includegraphics[width=0.9\linewidth]{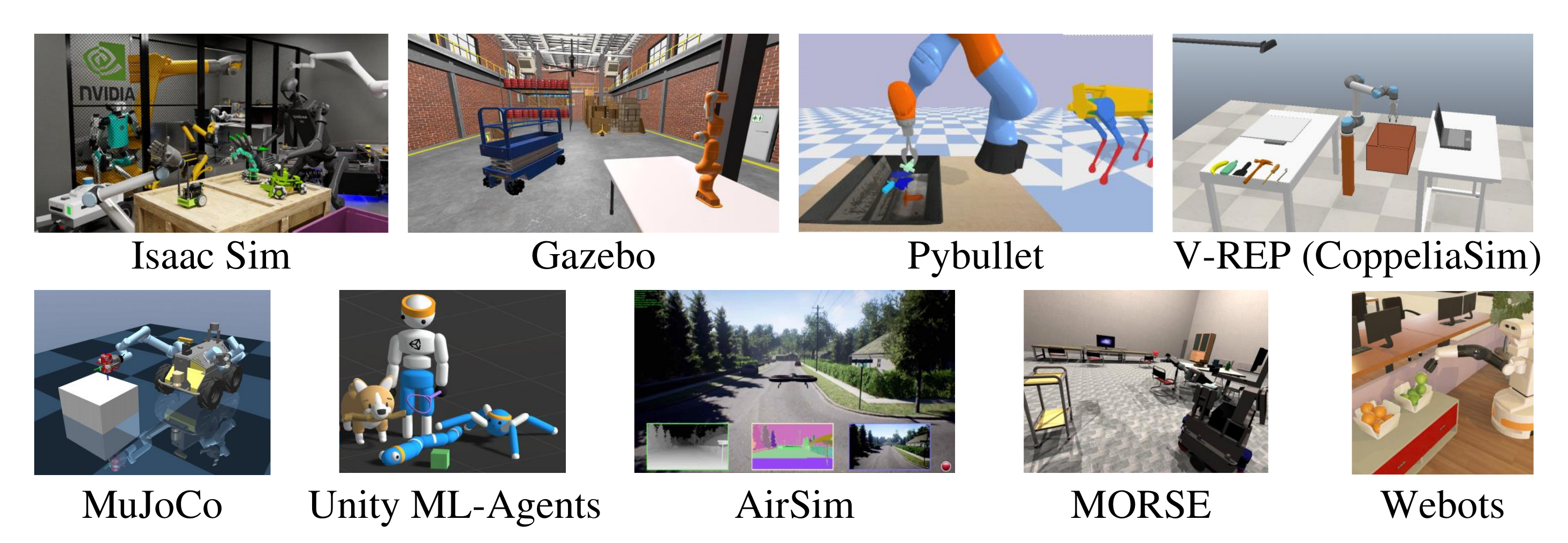}
        \vspace{-15pt}
    \caption{The schematic diagram of active visual perception. Visual SLAM and 3D Scene Understanding provide the foundation for passive visual perception, while active exploration provides activeness to the passive perception system. These elements works collaboratively for the active visual perception system.}
    \vspace{-15pt}
    \label{fig:Active-Visual-Perception}
\end{figure}

Active visual perception systems require fundamental capabilities such as state estimation, scene perception, and environment exploration. As shown in Fig. \ref{fig:Active-Visual-Perception}, these capabilities have been extensively studied within the domains of Visual Simultaneous Localization and Mapping (vSLAM)\cite{mokssit2023deep,chen2022semantic}, 3D Scene Understanding\cite{vinodkumar2023survey}, and Active Exploration\cite{hu2023toward}. These research areas contribute to developing robust active visual perception systems, facilitating improved environmental interaction and navigation in complex, dynamic settings. We briefly introduce these three components and summarize the methods mentioned in each part in Table \ref{tab:Active-Visual-Perception-methods}.

\subsubsection{Visual Simultaneous \red{Localization} and Mapping}
Simultaneous Localization and Mapping (SLAM) aims to determine a robot's position within an unknown environment while simultaneously constructing a map of the environment \cite{durrant2006simultaneous}. Range-based SLAM \cite{zhang20202d,ruan2020gp} relies on rangefinders, such as laser scanners, radar, and sonar, to generate point cloud representations. However, this approach is costly and provides limited environmental information. In contrast, Visual SLAM (vSLAM) \cite{mokssit2023deep,chen2022semantic} employs on-board cameras to capture image frames and build environmental representations. Its advantages include low hardware costs, high accuracy in small-scale scenarios, and the ability to capture rich environmental details. Classical vSLAM can be broadly categorized into Traditional vSLAM and Semantic vSLAM \cite{chen2022semantic}. Traditional vSLAM uses image data and multi-view geometry to estimate a robot's pose and construct low-level maps (e.g., sparse, semi-dense, or dense point clouds) through methods like filter-based approaches (e.g., MonoSLAM \cite{davison2007monoslam}), keyframe-based methods (e.g., ORB-SLAM \cite{mur2015orb}), and direct tracking techniques (e.g., LSD-SLAM \cite{engel2014lsd}). However, low-level maps do not directly correspond to objects, making them challenging for robots to interpret and utilize. Semantic vSLAM addresses this limitation by integrating semantic information, enhancing robots' ability to perceive and navigate unexplored environments.

\subsubsection{3D Scene Understanding}
3D scene understanding \cite{luo2025dspnet} aims to distinguish objects' semantics, identify their locations, and infer the geometric attributes from 3D scene data \cite{wei20253daffordsplat}, which is fundamental in autonomous driving\cite{mittal2020attngrounder}, robot navigation\cite{wang2019reinforced}, and human-computer interaction\cite{bermejo2021exploring} etc. A scene may be recorded as 3D point clouds using 3D scanning tools like LiDAR or RGB-D sensors. Unlike images, point clouds are sparse, disordered, and irregular. Recent advances in deep learning for 3D scene understanding can be categorized into projection-based, voxel-based, and point-based methods. Concretely, projection-based methods (e.g., MV3D\cite{chen2017multi}, PointPillars\cite{lang2019pointpillars}, MVCNN\cite{su2015multi} ) project 3D points onto various image planes and employ 2D CNN-based backbones for feature extraction. Voxel-based methods convert point clouds into regular voxel grids to facilitate 3D convolution operations (e.g., VoxNet\cite{maturana2015voxnet}, SSCNet\cite{song2017semantic}), and some works improve their efficiency through sparse convolution (e.g., MinkowskiNet\cite{choy20194d}, SSCNs\cite{graham20183d}, Embodiedscan\cite{wang2024embodiedscan}). In contrast, point-based methods process point clouds directly (e.g., PointNet\cite{qi2017pointnet}, PointNet++\cite{qi2017pointnet++}, PointMLP\cite{ma2022rethinking}). Recently, to achieve model scalability, Transformers-based (e.g., PointTransformer\cite{zhao2021point}, Swin3d\cite{yang2023swin3d}, PT2\cite{wu2022point}, 
3D-VisTA\cite{zhu20233d}, LEO\cite{huang2023embodied}, PQ3D\cite{zhu2024unifying}
) and Mamba-based (e.g., PointMamba\cite{liang2024pointmamba},
Mamba3D\cite{han2024mamba3d}) architectures have emerged. Notably, PQ3D\cite{zhu2024unifying} enhances scene understanding by seamlessly integrating features from point clouds, multi-view images, and voxels.


\subsubsection{Active Exploration}
The 3D scene understanding methods allow robots to passively perceive the environment, with static information acquisition and decision-making regardless of scene changes. Thus, while passive perception is essential, it must be complemented by active exploration, enabling robots to dynamically interact with and perceive their surroundings. The relationship between them is shown in Fig. \ref{fig:Active-Visual-Perception}. Current methods addressing active perception focus on interacting with the environment\cite{pinto2016curious,tatiya2023transferring} or by changing the viewing direction to obtain more visual information\cite{jayaraman2018learning,jin2023neu,hu2023off, fan2024evidential}.

For example, Pinto et al.\cite{pinto2016curious} proposed a curious robot that learns visual representations through physical interaction with the environment rather than relying solely on dataset category labels. To address the challenge of interactive object perception across robots with varying morphologies, Tatiya et al.\cite{tatiya2023transferring} proposed a multi-stage projection framework that transfers implicit knowledge through learned exploratory interactions.  Recognizing the challenge of autonomously capturing informative observations, Jayaraman et al. \cite{jayaraman2018learning} proposed a reinforcement learning method where an agent learns to actively acquire informative visual observations by reducing its uncertainty about unobserved parts of its environment. NeU-NBV\cite{jin2023neu} introduced a mapless planning framework that iteratively positions an RGB camera to capture the most informative images of an unknown scene. Hu et al.\cite{hu2023off} developed a robot exploration algorithm that predicts the value of future states using a state value function. To address the issue of accidental input in open-world environments, Fan et al.\cite{fan2024evidential} treated active recognition as a sequential evidence-gathering process, providing step-by-step uncertainty quantification and reliable prediction under evidence combination theory.



\subsection{Visual Language Navigation}

\begin{figure}[!t]
    \centering
\includegraphics[width=0.78\linewidth]{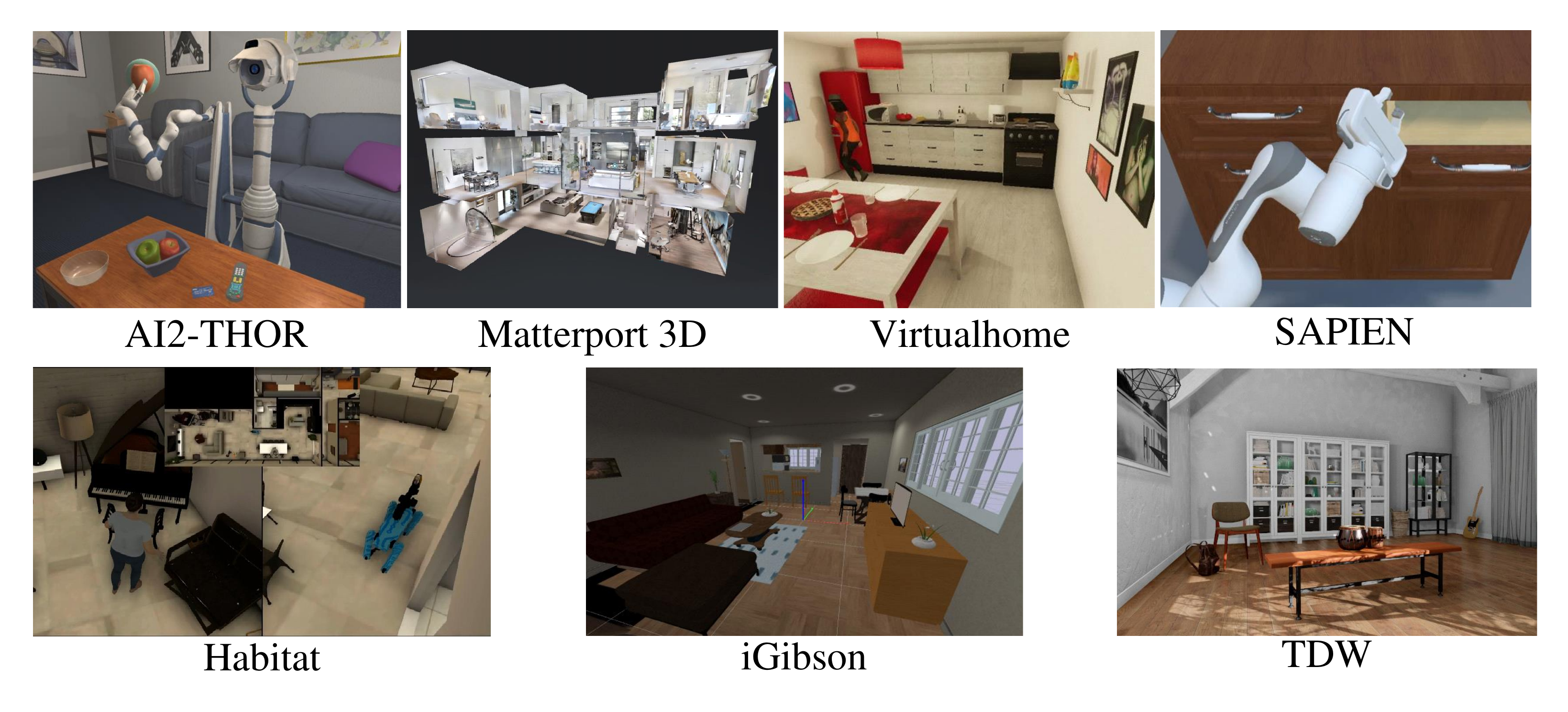}
    \vspace{-10pt}
    \caption{(a) Overview of VLN. The embodied agent communicates with humans through natural language. Humans issue instructions to the embodied agent, who completes tasks such as planning and dialog. Subsequently, through collaborative cooperation or the embodied agent’s independent actions, actions are made in interactive or non-interactive environments based on visual observations and instructions, (b) Different tasks of VLN.}
        \vspace{-15pt}
    \label{fig:VLN}
\end{figure}

Visual Language Navigation (VLN) is an essential task, aiming at navigating in unseen environments following linguistic instructions. VLN requires robots to understand complex and diverse visual observations and meanwhile interpret instructions at different granularities. The input typically consists of two parts: visual information and natural language instructions. The visual information can either be a video of past trajectories or a set of historical-current observation images. The natural language instructions include the target that the agent needs to reach or the task that the agent is expected to complete. The agent must use the above information to select one or a series of actions from a list of candidates to fulfill the requirements of the natural language instructions. This process could be represented as $
Action = \mathcal{M}(O, H, I) $, where $Action$ is the chosen action or a list of action candidates, $O$ is the current observation, $H$ is the historical information, and $I$ is the natural language instruction.

\begin{table}[!t]\tiny
\caption{Comparison of Different VLN Datasets. \textbf{M3D}: Matterport3D, \textbf{AT}: AI2-THOR, \textbf{OG}: OmniGibson, \textbf{I}: Indoor, \textbf{D}: Discrete, \textbf{O}: Outdoor, \textbf{C}: Continuous, \textbf{SbS}: Step-by-step instructions, \textbf{DGN}: Described goal navigation, \textbf{DDN}: Demand-driven navigation, \textbf{NwI}: Navigation with interaction, \textbf{LSNwI}: Long-span navigation with interaction, \textbf{D\&O}: Dialog and oracle}
    \vspace{-8pt}
\centering\scriptsize
\begin{tabular}{c|c|>{\centering\arraybackslash}p{0.8cm}|>{\centering\arraybackslash}p{1.3cm}|c|c}
\hline
Dataset        & Year    & Simulator     & Environment      & Feature       & Size    \\ \hline
R2R \cite{R2R}  & 2018     & M3D& I, D& SbS& 21,567  \\
R4R \cite{R4R}   &  2019    & M3D& I, D& SbS& 200,000+ \\
VLN-CE \cite{VLN-CE}  & 2020  & Habitat    & I, C& SbS& -   \\
TOUCHDOWN \cite{TOUCHDOWN}  & 2019   & -    & O, D& SbS& 9,326 \\
REVERIE \cite{REVERIE} & 2020  & M3D& I, D& DGN& 21,702 \\
SOON \cite{SOON}  &  2021   & M3D& I, D& DGN& 3,848  \\
DDN \cite{wang2023find} &   2023    &   AT&   I, C&   DDN&   30,000+\\
ALFRED \cite{ALFRED}   & 2020    & AT& I, C& NwI& 25,743     \\
OVMM \cite{OVMM}   &  2023     & Habitat     & I, C& NwI& 7,892       \\
BEHAVIOR-1K\cite{BEHAVIOR-1K} & 2023  & OG& I, C& LSNwI& 1,000     \\
CVDN \cite{CVDN}   & 2020    & M3D& I, D& D\&O& 2,050     \\
DialFRED \cite{DialFRED}   & 2022    & AT& I, C& D\&O& 53,000     \\ \hline
\end{tabular}
\vspace{-15pt}
\label{table:VLN dataset}
\end{table}


\subsubsection{Datasets}

In VLN, natural language instructions can be a series of detailed action descriptions, a fully described goal, or just a roughly described task, even only the demands of human. The tasks that embodied agents need to complete maybe just a single navigation, or navigation with interaction, or multiple navigation tasks that need to be completed in sequence. These differences bring different challenges to VLN, and many different datasets have been built. Based on these differences, we introduce some important VLN datasets.

\textbf{Room to Room (R2R)}\cite{R2R} is a VLN dataset based on Matterport3D. In R2R, embodied agents navigate according to step-by-step instructions, choosing the next adjacent navigation graph node to advance based on visual observations until they reach the target location. \textbf{Room-for-Room} \cite{R4R} extends the paths in R2R to longer trajectories, which requires stronger long-distance instruction and history alignment capabilities of embodied agents. \textbf{VLN-CE} \cite{VLN-CE} extends R2R and R4R to continuous environments, embodied agents can move freely in the scene. Different from the above datasets based on indoor scenes, the \textbf{TOUCHDOWN} dataset \cite{TOUCHDOWN} is created based on Google Street View. In TOUCHDOWN, embodied agents follow instructions to navigate in the street view rendering simulation of New York City to find the specified object. Similar to R2R, the \textbf{REVERIE} dataset \cite{REVERIE} is also built based on the Matterport3D simulator. REVERIE requires embodied agents to accurately locate the distant invisible target object specified by concise, human-annotated high-level natural language instructions. In \textbf{SOON} \cite{SOON}, agents receive a long and complex instruction from coarse to fine to find the target object in the 3D environment. During navigation, agents first search a larger area, and then gradually narrow the search range according to the visual scene and instructions. \textbf{DDN} \cite{wang2023find} moves a step further beyond these datasets, only providing human demands without specifying explicit objects. The agent needs to navigate through the scene to find objects.

\textbf{ALFRED} dataset \cite{ALFRED} is based on the AI2-THOR simulator. In ALFRED, embodied agents need to understand environmental observations and complete household tasks in an interactive environment according to coarse-grained and fine-grained instructions. The task in \textbf{OVMM} \cite{OVMM} dataset is to pick any object in any unseen environment and place it in a specified location. OVMM provides a simulation based on Habitat and a framework for implementation in the real world. \textbf{Behavior-1K} dataset \cite{BEHAVIOR-1K} is based on human needs, comprising 1,000 long-sequence, complex, skill-dependent daily tasks. Agents need to complete long-span navigation-interaction tasks which contain thousands of low-level action steps based on visual information and language instructions. These complex tasks requires strong capabilities of understanding and memory. \textbf{CVDN} \cite{CVDN} requires embodied agents to navigate to the target based on dialogue history, and ask questions for help to decide the next action when uncertain. \textbf{DialFRED} \cite{DialFRED}, an extension of ALFRED, allows agents to ask questions during the navigation and interaction process to get help. These datasets introduce additional oracles, and embodied agents need to obtain more information beneficial to navigation by asking questions.

\subsubsection{Method}

VLN has made great strides recently with the astonishing performance of LLMs, the direction and focus of VLN have been profoundly influenced. Nevertheless, the VLN methods can be divided into two directions: \textbf{Memory-Understanding Based} and \textbf{Future-Prediction Based}.

Memory-Understanding based methods focus on the perception and understanding of the environment, as well as model design based on historical observations or trajectories, which is a method based on past learning. Future-Prediction based methods pay more attention to modeling, predicting, and understanding the future state, which is a method for future learning. Since VLN can be regarded as a partially observable Markov decision process, where future observations depend on the current environment and actions of the intelligent agent, historical information has important significance for navigation decisions, especially long-span navigation decisions, hence Memory-Understanding based methods have always been the mainstream of VLN. However, Future-Prediction based methods still have important significance. Its essential understanding of the environment has great value in VLN in continuous environments, especially with the rise of the concept of world model, Future-Prediction based methods are receiving more and more attention from researchers.
\begin{table}[t]
\caption{Comparison of VLN methods.}
    \vspace{-8pt}
\centering\scriptsize
\begin{tabular}{c|c|c|c}
\hline
Method  & Model      & Year      & Feature       \\ \hline
&   LVERG   \cite{10.5555/3495724.3496368}  &   2020    &   Graph Learning\\
&   CMG \cite{Zhang2020LanguageGuidedNV}    &   2020    &   Adversarial Learning\\
&  RCM \cite{8986691}  &  2021   &    Reinforcement learning\\
&   FILM \cite{min2022film} & 2022    & Semantic Map\\
\textbf{Memory-}&   LM-Nav \cite{Shah2022LMNavRN}   &   2022    &   Graph Learning\\
\textbf{Understanding}&   HOP \cite{9880046}  &   2022    &   History Modeling \\
\textbf{Based}&   NaviLLM \cite{zheng2023learning}    &   2024    & Large Model\\
&  FSTT \cite{Gao2024Fast} &   2024    &   Test-Time Augmentation\\
&   DiscussNav \cite{long2023discuss}   &   2024    &   Large Model\\
&   GOAT \cite{wang2024learning}    &   2024    &   Causal Learning\\
&   VER \cite{liu2024volumetric}    &   2024    &   Environment Encoder\\
&   NaVid \cite{Zhang2024NaVidVV}   &   2024    &   Large Model
\\ \hline
&   LookBY \cite{Wang2018LookBY}    &  2018 &  Reinforcement Learning\\
\textbf{Future-}&   NvEM \cite{an2021neighbor}    &   2021    &   Environment Encoder\\
\textbf{Prediction}&   BGBL \cite{9879946}  &   2022    &   Graph Learning\\
\textbf{Based}&   Mic \cite{Qiao_2023_MiC}    &   2023    &   Large Model\\
&   HNR \cite{wang2024lookahead}    &   2024    &   Environment Encoder\\
&   ETPNav \cite{an2024etpnav}  &   2024    &   Graph Learning
\\ \hline
\multirow{2}{*}{\textbf{Others}}
&   MCR-Agent \cite{10.1609/aaai.v37i1.25094}   &   2023    &   Multi-Level Model\\
&   OVLM \cite{Xu2023VisionAL}   &   2023    &   Large Model
\\ \hline
\end{tabular}
\vspace{-15pt}
\label{table:VLN method}
\end{table}

\textbf{Memory-Understanding based. }Graph-based learning is an essential part of the memory-understanding based method. It usually represents the navigation process in the form of a graph, where the information obtained by the agent at each time step is encoded as nodes of the graph. The agent obtains global or partial navigation graph information as a representation of the historical trajectory. LVERG \cite{10.5555/3495724.3496368} encoded the language information and visual information of each node separately, design a new language and visual entity relationship graph to model the inter-modal relationship between text and vision, and the intra-modal relationship between visual entities. LM-Nav \cite{Shah2022LMNavRN} used a goal-conditioned distance function to infer connections between original observation sets and construct a navigation graph, and extracted landmarks from the instructions through a LLM. Although HOP \cite{9880046} is not based on graph learning, it requires to model time-ordered information at different granularities, thereby achieving a deep understanding of historical trajectories and memories.

The navigation graph discretizes the environment, but concurrently understanding and encoding the environment is also important. FILM \cite{min2022film} used RGB-D observations and semantic segmentation to gradually build a semantic map from 3D voxels during the navigation. VER \cite{liu2024volumetric} quantified the physical world into structured 3D units through 2D-3D sampling, providing fine-grained geometric details and semantics.

Different learning schemes explore how to utilize historical trajectories and memories better. Through adversarial learning, CMG \cite{Zhang2020LanguageGuidedNV} alternated between imitation learning and exploration encouragement schemes, effectively strengthening the understanding of instructions and historical trajectories, shortening the difference between training and inference. GOAT \cite{wang2024learning} directly trained unbiased models through Backdoor Adjustment Causal Learning (BACL) and Frontdoor Adjustment Causal Learning (FACL), conducted contrastive learning with vision, navigation history, and their combination to instructions, enabling the agent to make fuller use of information. The enhanced cross-modal matching method proposed by RCM \cite{8986691} used goal-oriented external rewards and instruction-oriented internal rewards to perform cross-modal grounding globally and locally and learns from its own historical good decisions through self-supervised imitation learning. FSTT \cite{Gao2024Fast} introduced TTA into VLN and optimizes the model in terms of gradients and model parameters at two scales of time steps and tasks, effectively improving model performance.

The specific application of large models in Memory-Understanding based methods is to understand the representation of historical memory and to understand the environment and tasks based on its extensive world knowledge. NaviLLM \cite{zheng2023learning} integrated the historical observation sequence into the embedding space through the visual encoder, inputs the multi-modal information of the fusion encoding into the LLM and fine-tunes it, reaching the state-of-the-art on multiple benchmarks. NaVid \cite{Zhang2024NaVidVV} \red{made} improvements in the encoding of historical information, achieves different degrees of information retention on historical observations and current observations through different degrees of pooling. LH-VLN \cite{song2024towards}  proposed the NavGen platform, the long-horizon navigation benchmark, and the Multi-Granularity Dynamic Memory (MGDM) module to enhance task evaluation and model adaptability in dynamic environments.

\textbf{Future-Prediction Based. }Graph-based learning is also widely used in Future-Prediction based methods. BGBL \cite{9879946} and ETPNav \cite{an2024etpnav} used a similar method to design a waypoint predictor that can predict movable path points in a continuous environment based on the observation of the current navigation graph node. They aim to migrate complex navigation in a continuous environment to node-to-node navigation in a discrete environment, thereby bridging the performance gap from discrete environments to continuous environments.

Improving the understanding and perception of the future environment through environmental encoding is also one of the research directions for predicting and exploring the future. NvEM \cite{an2021neighbor} used a theme module and a reference module to perform fusion encoding of neighbor views from the global and local perspectives. This is actually an understanding and learning of future observations. HNR \cite{wang2024lookahead} used a large-scale pre-trained hierarchical neural radiation representation model to directly predict the visual representation of the future environment rather than pixel-level images using three-dimensional feature space encoding, and builds a navigable future path tree based on the representation of the future environment. They predict the future environment from different levels, providing effective references for navigation decisions.

Some reinforcement learning methods are also applied to predict and explore future states. LookBY \cite{Wang2018LookBY} employed reinforcement prediction to enable the prediction module to imitate the world and forecast future states and rewards. This allows the agent to directly map ``current observations" and ``predictions of future observations" to actions, achieving state-of-the-art performance at the time. The rich world knowledge and zero-shot performance of large models provide many possibilities for Future-Prediction based methods. MiC \cite{Qiao_2023_MiC} required the LLM to directly predict the target and its possible location from the instructions and provides navigation instructions through the description of scene perception. This method requires LLMs to fully exert its ‘imagination’ and build an imagined scene through prompts.

In addition, there are some methods that both learn from the past and for the future. MCR-Agent \cite{10.1609/aaai.v37i1.25094} designed a three-layer action strategy, which requires the model to predict the target from the instructions, predict the pixel-level mask for the target to be interact, and learn from the previous navigation decision; OVLM \cite{Xu2023VisionAL} required the LLMs to predict the corresponding operations and landmark sequences for the instructions. During the navigation process, the visual language map will be continuously updated and maintained, and the operations will be linked to the waypoints on the map.


\section{Embodied Interaction} 

Embodied interaction refer to scenarios where agents interact with humans and the environment in physical or simulated space. The typical embodied interaction tasks are Embodied Question Answering (EQA) and embodied grasping.

\begin{figure}[!t]
    \centering
    \includegraphics[width=0.95\linewidth]{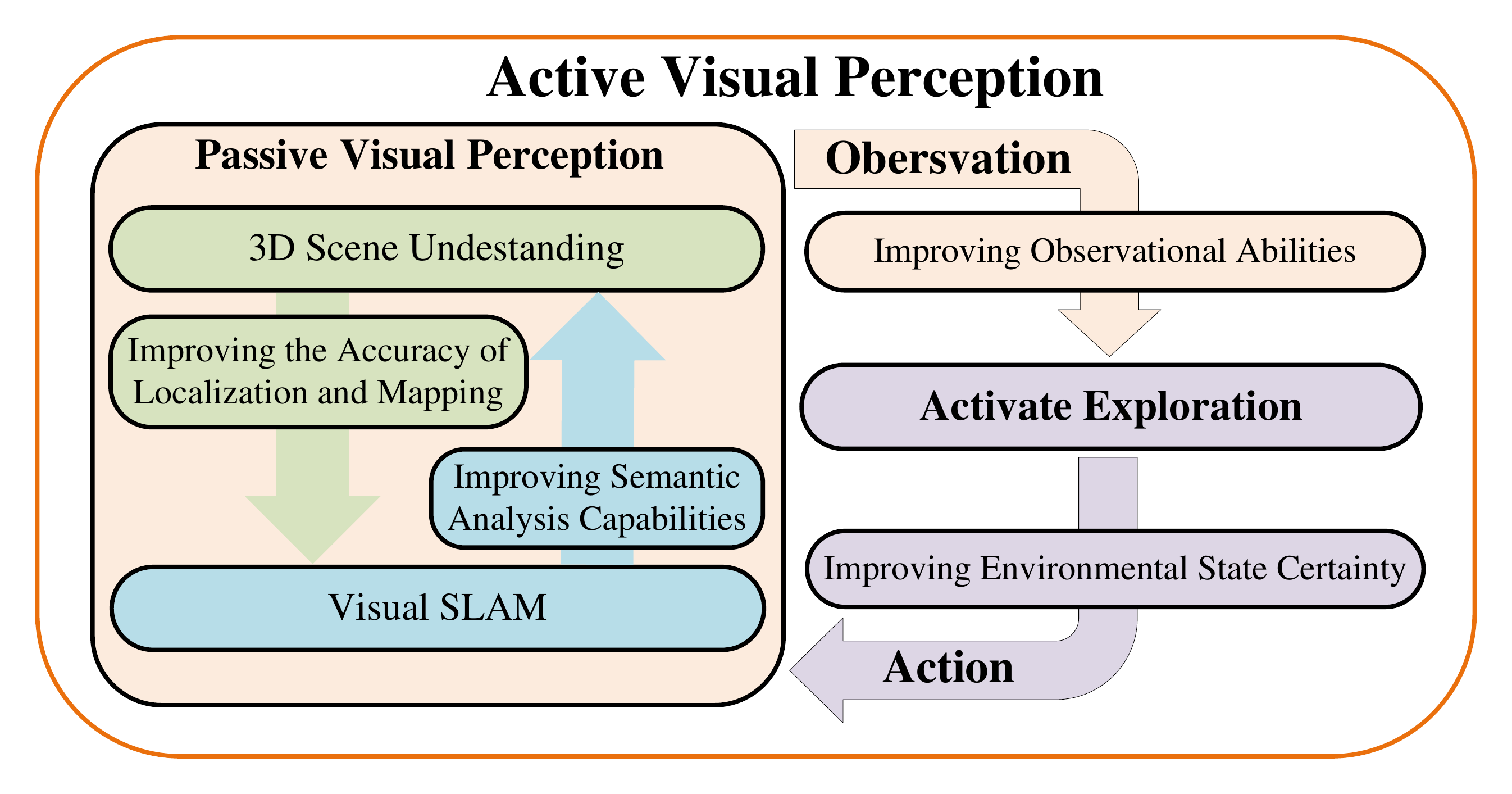}
    \vspace{-15pt}
    \caption{The gray box displays the scenes an agent observes during exploration. The other boxes show various types of question answering tasks. Except for the task of answering questions based on episodic memory, the agent ceases exploration once it has gathered sufficient information to answer the question.}
        \vspace{-10pt}
    \label{fig:EQA}
\end{figure}

\subsection{Embodied Question Answering}

For EQA task, the agent needs to explore the environment from a first-person perspective to gather information necessary to answer the given questions. An agent with autonomous exploration and decision-making capabilities must not only consider which actions to take to explore the environment but also determine when to stop exploring to answer questions. Existing works focus on different types of questions, some of which are shown in Fig. \ref{fig:EQA}. In this section, we will introduce the existing datasets, discuss the related methods, describe the metrics used to evaluate model performance, and address the remaining limitations of this task.

\subsubsection{Datasets} We briefly introduce several embodied question answering datasets, which are summarized in Table \ref{table:dataset}.

\begin{table*}[!t]\tiny
\caption{Comparison of Different EQA Datasets.}
\vspace{-8pt}
\centering\scriptsize
\begin{tabular}{c|c|c|c|c|c|c|c}
\hline
Dataset        & Year   & Type                            & Data Sources  & Simulator                  & Query Creation & Answer       & Size    \\ \hline
EQA v1\cite{das2018embodied}  & 2018  & Active EQA                      & SUNCG         & House3D                    & Rule-Based     & open-ended   & 5,000+  \\
MT-EQA\cite{yu2019multi}   &  2019  & Active EQA                      & SUNCG         & House3D                    & Rule-Based     & open-ended   & 19,000+ \\
MP3D-EQA\cite{wijmans2019embodied}  & 2019 & Active EQA                      & MP3D          & Simulator based on MINOS & Rule-Based     & open-ended   & 1,136   \\
IQUAD V1\cite{gordon2018iqa}  & 2018 & Interactive EQA                 & -             & AI2THOR                    & Rule-Based     & multi-choice & 75,000+ \\
VideoNavQA\cite{cangea2019videonavqa} & 2019 & Episodic Memory EQA             & SUNCG         & House3D                    & Rule-Based     & open-ended   & 101,000 \\
SQA3D\cite{ma2022sqa3d} & 2022 & QA only & ScanNet & - & Manual & multi-choice & 33,400 \\
K-EQA\cite{tan2023knowledge}  &  2023  & Active EQA                      & -             & AI2THOR                    & Rule-Based     & open-ended   & 60,000  \\
OpenEQA\cite{majumdar2024openeqa}  & 2024 & Active EQA, Episodic Memory EQA & ScanNet, HM3D & Habitat                    & Manual         & open-ended   & 1,600+  \\
HM-EQA\cite{ren2024explore}   & 2024  & Active EQA                      & HM3D             & Habitat                    & VLM            & multi-choice & 500     \\
S-EQA\cite{dorbala2024s}   &  2024  & Active EQA                      & -          & VirtualHome                & LLM            & binary       & -       \\
\red{EXPRESS-Bench\cite{EXPRESSBench} }  &  2025  & Exploration-aware EQA                      & HM3D         &  Habitat                & VLM            & open-ended        & 2,044      \\
\hline
\end{tabular}
\vspace{-15pt}
\label{table:dataset}
\end{table*}

\textbf{EQA v1}\cite{das2018embodied} is the first dataset designed for EQA. Built on synthetic 3D indoor scenes from the SUNCG dataset\cite{song2017semantic} within the House3D\cite{wu2018building} simulator, EQA v1 comprises four types of questions: location, color, color\_room, and preposition. Similar to EQA v1, \textbf{MT-EQA}\cite{yu2019multi} is built in House3D using SUNCG by executing functional programs consisting of some basic operations. However, it further extends the single-object question answering task to a multi-object setting. Six types of questions are designed, involving the comparison of color, distance, and size between multiple objects. \textbf{MP3D-EQA}\cite{wijmans2019embodied} is built on a simulator developed based on MINOS\cite{savva2017minos} using the Matterport3D dataset\cite{chang2017matterport3d}, expanding the question-answering task to a realistic 3D environment. Referring to EQA v1, MP3D-EQA utilizes three types of templates: location, color, and color\_room, generating a total of 1,136 questions in 83 home environments. \textbf{IQUAD V1}\cite{gordon2018iqa} is built upon AI2-THOR and consists of three types of questions: existence, counting, and spatial relationships. Unlike other datasets, answering IQUAD V1 questions requires the agent to have a good understanding of affordances and interact with the dynamic environment. \textbf{VideoNavQA}\cite{cangea2019videonavqa} decouples the visual reasoning from the navigation aspect of the EQA problem. In this task, the agent accesses videos corresponding to exploration trajectories with sufficient information to answer questions.  \textbf{SQA3D}\cite{ma2022sqa3d} simplifies protocol (QA only) while still preserving the function of benchmarking embodied scene understanding, enabling more complex, knowledge-intensive questions and a much larger scale of data collection.

Unlike previous datasets that explicitly specify target objects in questions, \textbf{K-EQA}\cite{tan2023knowledge} features complex questions with logical clauses and knowledge-related phrases, requiring prior knowledge to answer.  \textbf{OpenEQA}\cite{majumdar2024openeqa} is the first open-vocabulary dataset for EQA, supporting both episodic memory and active exploration cases. The episodic memory EQA (EM-EQA) tasks involve an agent developing an understanding of the environment from its episodic memory to answer questions. In active EQA (A-EQA) tasks, the agent answers questions by taking exploratory actions to gather necessary information. Utilizing GPT4-V, \textbf{HM-EQA}\cite{ren2024explore} is constructed in the Habitat simulator using HM3D. It includes 500 questions across 267 different scenes, which can be roughly categorized into identification, counting, existence, status, and location. \textbf{S-EQA}\cite{dorbala2024s} leverages GPT-4 in VirtualHome for data generation and employs cosine similarity calculations to decide whether to retain the generated data, thereby enhancing dataset diversity. In S-EQA, answering questions requires the assessment of a collection of consensus objects and states to reach an existential ``Yes/No" answer. \red{ \textbf{EXPRESS-Bench} \cite{EXPRESSBench} is the largest exploration-aware EQA dataset that consists of 777 exploration trajectories and 2,044 samples. It also introduces novel evaluation metrics to ensure faithful assessment.}

\subsubsection{Methods} The embodied question answering task mainly involves navigation and question-answering subtasks, with implementation methods broadly categorized into two types: neural network-based and LLMs/VLMs-based.

\textbf{Neural Network Methods.} In early work, researchers mainly addressed the embodied question answering task by building deep neural networks. They trained and fine-tuned these models using techniques such as imitation learning and reinforcement learning to improve performance.

The EQA task was first proposed by Das et al.\cite{das2018embodied}. In their work, the agent consists of four main modules: vision, language, navigation, and answering. These modules are primarily constructed using traditional neural building blocks:  Convolutional Neural Networks (CNNs) and Recurrent Neural Networks (RNNs). Some subsequent works\cite{wu2020revisiting} retained modules like the question answering module proposed by Das et al.\cite{das2018embodied} and improved the model. Additionally, Wu et al.\cite{wu2020revisiting} proposed integrating the navigation and QA modules into a unified SGD training pipeline for joint training, thereby avoiding employing deep reinforcement learning to simultaneously train the separately trained navigation and question answering modules.
From the perspective of task singularity, several works\cite{tan2020multi} expanded the task to include multiple objectives and multi-agent, respectively, making it necessary for the model to store and integrate the information obtained by the agent's exploration through methods such as feature extraction and scene reconstruction.
Considering the interaction between the agent and the dynamic environment, Gordon et al.\cite{gordon2018iqa} introduced the Hierarchical Interactive Memory Network. There is also a limitation in previous works where agents are unable to use external knowledge to answer complex questions and lack knowledge of the explored parts of the scene. To address this, Tan et al.\cite{tan2023knowledge} leveraged the neural program synthesis method and the table converted from the knowledge and 3D scene graphs, allowing the action planner to access object-related information. Additionally, an approach based on Monte Carlo Tree Search (MCTS) is used to determine the next location for the agent.

\begin{figure*}[!t]
    \centering
    \includegraphics[width=0.88\linewidth]{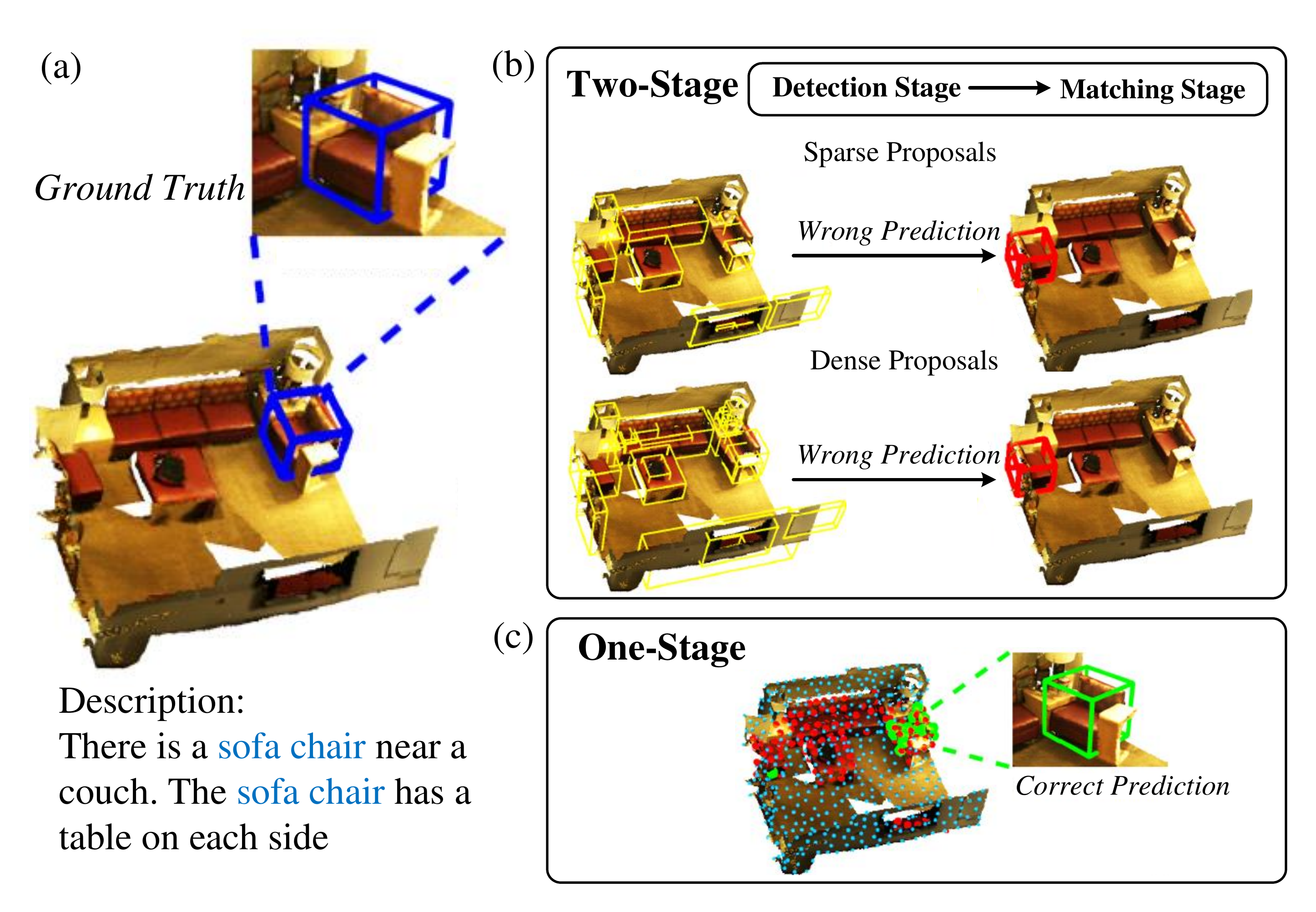}
    \vspace{-10pt}
    \caption{The overview of the embodied grasping task. (a) demonstrates examples of language-guided grasping for different types of tasks, (b) provides an overview of human-agent-object interaction, (c) shows Google Scholar search results for topics of ``Language-guided Grasping”. }
        \vspace{-10pt}
    \label{fig:language-guided grasping}
\end{figure*}

\textbf{LLMs/VLMs Methods.}
Majumdar et al.\cite{majumdar2024openeqa} used LLMs and VLMs for episodic memory EQA (EM-EQA) and Active EQA (A-EQA) tasks. For EM-EQA task, they considered Blind LLMs, Socratic LLMs with language descriptions of the episodic memory, Socratic LLMs with descriptions of the constructed scene graph, and VLMs processing multiple scene frames. The A-EQA task extended EM-EQA methods with frontier-based exploration (FBE)\cite{yamauchi1997frontier} for problem-independent environment exploration.
Some works\cite{ren2024explore, sakamoto2024map} employed frontier-based exploration method to identify areas for subsequent exploration and to build semantic maps. They ended the exploration early utilizing conformal prediction or image-text matching to avoid over-exploration.
Patel et al.\cite{patel2024embodied} emphasized the question answering aspect of the task. They leveraged multiple LLM-based agents to explore the environment and enable them to independently answer questions with ``yes" or ``no" answers. These individual responses are utilized to train a Central Answer Model, to aggregate responses and generate robust answers.


\subsubsection{Limitations} a) Dataset: Constructing datasets requires substantial manpower and resources. Additionally, there are still few large-scale datasets, and the metrics for evaluating model performance vary across different datasets, complicating the testing and comparison of performance, b) Model: Despite the advancements brought by LLMs, the performance of these models still lags significantly behind human levels. Future work may focus more on effectively storing environmental information explored by agents and guiding them to plan actions based on environmental memory and questions, while also enhancing the interpretability of their actions.

\subsection{Embodied Grasping} 

Embodied interaction includes not only question-answering but also performing tasks like grasping and placing objects based on human instructions. Embodied grasping combines traditional kinematic methods~\cite{liu2024multifingered, chen2021edge} with large models such as LLMs and vision-language models, enabling multi-sensory perception and reasoning for task execution. Figure \ref{fig:language-guided grasping} (b) shows an overview of human-agent-object interactions where embodied grasping is performed.

\begin{table*}[!t]\setlength{\tabcolsep}{3.5mm}
    \caption{Embodied grasping datasets.}
        \vspace{-8pt}
\centering\scriptsize
    \begin{tabular}{c|c|c|c|c|c|c|c|c|c} \hline
         Dataset&  Year&  Type &Modality &Grasp
Label&  Gripper Finger&Objects&  Grasps&  Scenes& Language\\ \hline
         Cornell \cite{jiang2011efficient} &  2011&  Real & RGB-D&Rect.&  2&  240&  8K&  Single& ×\\
         Jacquard \cite{depierre2018jacquard}&  2018&  Sim & RGB-D&Rect.&  2&  11K&  1.1M&  Single& ×\\
         6-DOF GraspNet \cite{mousavian20196}&  2019&  Sim & 3D&6D&  2&  206&  7.07M&  Single& ×\\
 ACRONYM \cite{eppner2021acronym}& 2021& Sim & 3D&6D& 2& 8872& 17.7M& Multi&×\\
 MultiGripperGrasp \cite{murrilo2024multigrippergrasp}& 2024& Sim & 3D&-& 2-5& 345& 30.4M& Single&×\\
 OCID-Grasp \cite{ainetter2021end}& 2021& Real & RGB-D&Rect.& 2& 89& 75K& Multi&×\\
 OCID-VLG \cite{tziafas2023language}& 2023& Real & RGB-D,3D&Rect.& 2& 89& 75K& Multi& $\surd$ \\
 ReasoingGrasp \cite{jin2024reasoning}& 2024& Real & RGB-D&6D& 2& 64& $~$99.3M& Multi& $\surd$ \\
 CapGrasp \cite{li2024semgrasp}& 2024&  Sim& 3D&-& 5& 1.8K& 50K& Single& $\surd$ \\\hline
    \end{tabular}
    \vspace{-15pt}
    \label{ Grasping Datasets}
\end{table*}

\subsubsection{Datasets}

 Recently, a substantial number of grasping datasets \cite{jiang2011efficient, depierre2018jacquard, mousavian20196, eppner2021acronym, murrilo2024multigrippergrasp} have been generated. These datasets typically contain annotated grasping data based on images (RGB, depth), point clouds, or 3D scenes. With the advent of MLMs and the application of foundational language models to robotic grasping, there is an urgent need for datasets that include linguistic text. Consequently, existing datasets have been extended or reconstructed to create semantic-grasping datasets \cite{tziafas2023language, jin2024reasoning, li2024semgrasp}. These datasets are instrumental in studying grasping models grounded in language, enabling agents to develop a broad understanding of semantics.

Traditional grasping datasets encompass data for both single objects \cite{jiang2011efficient} and cluttered scenes \cite{ainetter2021end}, providing stable grasp annotations (4-DOF or 6-DOF) that conform to kinematics for each object. These data can be collected from real desktop environments \cite{jiang2011efficient}, typically including RGB, depth, and point cloud data, or from virtual environments \cite{eppner2021acronym}, which include image data, point clouds, or scene models. While these datasets are useful for grasping models, they lack semantic information. To bridge this gap, these datasets have been augmented or extended with semantic expressions \cite{tziafas2023language, shridhar2022cliport}, thereby linking language, vision, and grasping. By incorporating semantic information, agents can better understand and execute grasping tasks. This enhancement allows for the development of more sophisticated and semantically aware grasping models, facilitating more intuitive and effective interaction with the environment. Table \ref{ Grasping Datasets} presents the datasets described above, including traditional grasping datasets and language-based grasping datasets.

\subsubsection{Language-guided grasping}

The concept of language-guided grasping \cite{tziafas2023language, shridhar2022cliport, jin2024reasoning}, which has evolved from this integration, combines MLMs to provide agents with the capability of semantic scene reasoning. This allows the agent to execute grasping operations based on implicit or explicit human instructions. Figure \ref{fig:language-guided grasping} (c) illustrates the publication trends in recent years on the topic of language-guided grasping. With the advancement of LLMs, researchers have shown increasing interest in this topic. Currently, grasping research is increasingly focused on open-world scenarios, emphasizing the open-set generalization \cite{shen2023F3RM} methods. By leveraging the generalization capabilities of MLMs, robots can perform grasping tasks in open-world environments with greater intelligence and efficiency.

In language-guided grasping, semantics can originate from explicit instructions \cite{shen2023F3RM, zheng2024gaussiangrasper} and implicit instructions \cite{jin2024reasoning, li2024semgrasp}. Explicit instructions clearly specify the category of the object to be grasped, such as a banana or an apple. Implicit instructions, however, require reasoning to identify the object or a part of the object to be grasped, involving spatial reasoning and logical reasoning.

Spatial reasoning \cite{tziafas2023language} refers to instructions that may include the spatial relationship of the object or part to be grasped, necessitating the inference of grasping posture based on the spatial relationships of objects within the scene. For example, ``Grasp the keyboard that is to the right of the brown kleenex box" involves understanding and inferring the spatial arrangement of objects. Logical reasoning \cite{jin2024reasoning}, on the other hand, involves instructions that may contain logical relationships requiring inference to discern human intent and subsequently grasp the target. For instance, ``I am thirsty, can you give me something to drink?" would prompt the agent to potentially hand over a glass of water or a bottle of a beverage. The agent must ensure that the liquid does not spill during the handover, thus generating a reasonable grasping posture.

In both cases, the integration of semantic understanding with spatial and logical reasoning enables the agent to perform complex grasping tasks effectively and accurately. Figure \ref{fig:language-guided grasping} (a) depicts various types of language-guided grasping tasks.

\subsubsection{End-to-End Approaches}

CLIPORT \cite{shridhar2022cliport} is a language-conditioned imitation learning agent that combines the vision-language pre-trained model CLIP with the Transporter Net to create an end-to-end dual-stream architecture for semantic understanding and grasp generation. It is trained using a large number of expert demonstration data collected from virtual environments, enabling the agent to perform semantically guided grasping. Based on the OCID dataset, CROG \cite{tziafas2023language} proposes a vision-language-grasping dataset and introduces a competitive end-to-end baseline. It leverages CLIP's visual foundation capabilities to learn grasp synthesis directly from image-text pairs. Reasoning Grasping  \cite{jin2024reasoning} introduces the first reasoning grasping benchmark dataset based on the GraspNet-1 Billion dataset and proposes an end-to-end reasoning grasping model. The model integrates multimodal LLMs with vision-based robotic grasping frameworks to generate grasps based on semantics and vision. SemGrasp \cite{li2024semgrasp} is a method for semantic-based grasp generation that incorporates semantic information into grasp representations to generate dexterous hand grasp postures. It introduces a discrete representation aligning grasp space with semantic space, enabling the generation of grasp postures according to language instructions.

\subsubsection{Modular Approaches}

F3RM \cite{shen2023F3RM} seeks to elevate CLIP's text-image priors into 3D space, using extracted features for language localization followed by grasp generation. It combines precise 3D geometry with rich semantics from 2D foundational models, utilizing features extracted from CLIP to specify objects for manipulation through free-text natural language. It demonstrates the ability to generalize to unseen expressions and new object categories. GaussianGrasper \cite{zheng2024gaussiangrasper} utilizes a 3D Gaussian field to achieve language-guided grasping tasks. The proposed methodology begins with the construction of a 3D Gaussian field, followed by feature distillation.  Subsequently, language-based localization is performed using the extracted features.  Finally, grasp pose generation is carried out based on a SOTA pre-trained grasping network \cite{fang2023anygrasp}. It integrates open-vocabulary semantics with precise geometry, enabling grasping based on language instructions.

These approaches advance language-guided grasping by using end-to-end and modular frameworks, enhancing robotic agents' ability to perform complex grasping tasks from natural language instructions. Embodied grasping improves robots' intelligence and utility in-home services and industrial manufacturing. However, current methods face limitations, including reliance on extensive data and poor generalization. Future research aims to enhance agent generality, enabling robots to understand complex semantics, grasp a variety of unseen objects, and tackle intricate tasks.

\begin{figure*}[!t]
    \centering
    \includegraphics[scale=0.36]{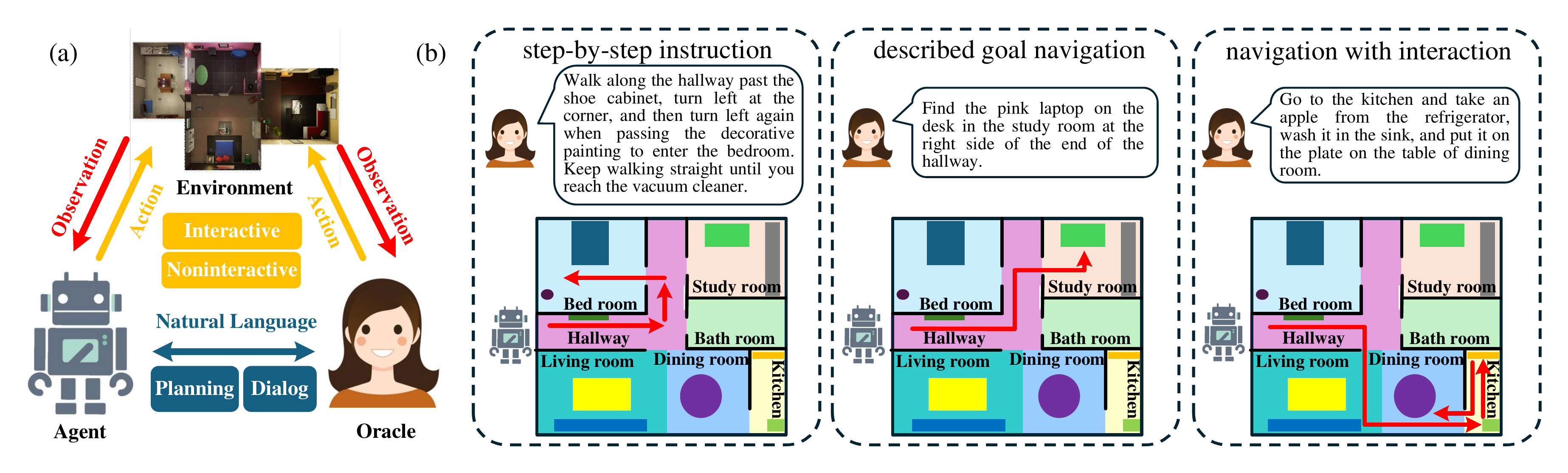}
        \vspace{-15pt}
    \caption{\red{The architecture of the embodied agent based on embodied multimodal foundation model, which consists of visual perception module, high-level task planning module, and low-level action planning module.}}
    \vspace{-15pt}
    \label{fig:Agent}
\end{figure*}

\section{Embodied Agent}


An agent is defined as an autonomous entity capable of perceiving its environment and acting to achieve specific objectives. Recent advancements in MLMs have further expanded the application of agents to practical scenarios. When these MLM-based agents are embodied in physical entities, they can effectively transfer their capabilities from virtual space to physical world, thereby becoming embodied agents~\cite{xi2023rise}.


To enable embodied agents to operate in the information-rich and complex real world, the embodied agents have been developed to show strong multimodal perception, interaction and planning capabilities, as shown in Fig.~\ref{fig:Agent}.
To complete a task, embodied agents typically involves the following process: 1) decomposing the abstract and complex task into specific subtasks, which is referred to as high-level Embodied Task Planning. 2) gradually implementing these subtasks by effectively utilizing Embodied Perception and Embodied Interaction models or leveraging the Foundation Model's policy function, named low-level Embodied Action Planning.
It is worth noting that task planning involves thinking before acting, and is therefore typically considered in cyber space. In contrast, action planning must account for effective interaction with the environment and feedback on this information to the task planner to adjust task planning. Thus, it is crucial for embodied agents to align and generalize their abilities from the cyber space to the physical world.


\subsection{Embodied Task Planning}
\label{sec:hlp}

Traditional embodied task planning methods are usually based on explicit rules and logical reasoning. For example, symbolic planning algorithms such as 
PDDL\cite{McDermott1998PDDLthePD}, and search algorithms like MCTS \cite{Metropolis1949TheMC} and A* \cite{Hart1968AFB}, are used to generate plans. However, these methods often rely on predefined rules, constraints, and heuristics that are rigid and may not adapt well to dynamic or unforeseen changes.
With the popularity of LLMs, many works have attempted to use LLMs for planning or to combine traditional methods with LLMs, leveraging the rich embedded world knowledge  for reasoning and planning without the need for handcrafted definitions, greatly enhancing the model's generalization capabilities.

\subsubsection{Planning utilizing the Emergent Capabilities of LLMs}

Before the scale-up of natural language models, task planners were similarly implemented by training models like BERT on embodied instruction datasets such as Alfred~\cite{shridhar2020alfred} and Alfworld~\cite{shridhar2020alfworld}, as demonstrated by FILM~\cite{min2021film}. However, this approach was limited by the examples in the training set and could not effectively align with the physical world.
Nowadays, thanks to the emergent capabilities of LLMs, they can decompose abstract tasks using their internal world knowledge and chain-of-thought reasoning, similar to how humans reason through task completion steps before acting. For example, Translated LM~\cite{huang2022language} and Inner Monologue~\cite{huang2023inner} can break down complex tasks into manageable steps and devise solutions using their internal logic and knowledge systems without additional training.
Similarly, the multi-agent collaboration framework ReAd \cite{zhang2024towards} efficiently self-refined plans via different prompts. Additionally, some approaches abstract past successful examples into a series of skills stored in a memory bank to consider during inference and improve planning success rates~\cite{sarchOpenEndedInstructableEmbodied2023, wangVoyagerOpenEndedEmbodied2023, Sharma2021SkillIA}.
%
%
Some works utilized code as the reasoning medium instead of natural language, where task planning is generated as code based on the available API library\cite{singhProgPromptGeneratingSituated2022,Liang2022CodeAP}. 
Furthermore, multi-turn reasoning can effectively correct potential hallucinations in task planning. For instance, Socratic Models \cite{zeng2023socratic} and Socratic Planner \cite{shin2024socraticplannerinquirybasedzeroshot} used Socratic questioning to derive reliable planning.

However,
during task planning, potential failures may occur during execution, often resulting from the planner not fully accounting for the complexity of the real environment and the difficulty of task execution~\cite{huang2023inner, Song2022LLMPlannerFG}. Due to a lack of visual information, planned subtasks may deviate from the actual scenario, leading to task failure. Therefore, integrating visual information into planning or replanning during execution is necessary. This approach can significantly enhance the accuracy and feasibility of task planning, better addressing the challenges of real-world environments.

\subsubsection{Planning utilizing the visual information from embodied perception model}

Based on the above discussion, it is important to integrate visual information into task planning (or replanning). In this process, object labels, locations, or descriptions provided by visual input can offer critical references for task decomposition and execution by LLMs. Through visual information, LLMs can more accurately identify target objects and obstacles in the current environment, thereby optimizing task steps or modifying subtask objectives. Some works use an object detector to query the objects present in the environment during task execution and feed this information back to the LLM, allowing it to modify unreasonable steps in the current plan~\cite{Song2022LLMPlannerFG, zeng2023socratic, wuEmbodiedTaskPlanning2023}. RoboGPT considers the different names of similar objects within the same task, further improving the feasibility of replanning~\cite{chen2023robogpt}. However, the information provided by labels is still too limited. Can further scene information be provided? SayPlan~\cite{Rana2023SayPlanGL} proposes using hierarchical 3D scene graphs to represent the environment, effectively mitigating the challenges of task planning in large, multi-floor, and multi-room settings. Similarly, ConceptGraphs~\cite{Gu2023ConceptGraphsO3} also adopts 3D scene graphs to provide environmental information to LLMs. Compared to SayPlan, it offers more detailed open-world object detection and presents task planning in a code-based format, which is more efficient and better suited to the demands of complex tasks.

However, limited visual information can result in an agent's inadequate understanding of its environment. While LLMs are provided with visual cues, they often fail to capture the environment's complexity and dynamic changes, leading to misunderstandings and task failures. For example, if a towel is locked in a bathroom cabinet, the agent might repeatedly search the bathroom without considering this possibility \cite{chen2023robogpt}. To address this, more robust algorithms must be developed to integrate multiple sensory data, enhancing the agent's environmental understanding. Additionally, leveraging historical data and contextual reasoning, even when visual information is limited, can aid the agent in making reasonable judgments and decisions. This approach of multimodal integration and context-based reasoning not only increases task execution success rates but also provides new perspectives for the advancement of embodied artificial intelligence.

\subsubsection{Planning utilizing the VLMs}

Compared to converting environmental information into text using external visual models, VLM models can capture visual details in latent space, particularly contextual information that is difficult to represent with object labels. VLMs can discern rules underlying visual phenomena; for instance, even if a towel is not visible in the environment, it can be inferred that the towel might be stored in a cabinet. This process essentially demonstrates how abstract visual features and structured textual features can be more effectively aligned in latent space. In EmbodiedGPT~\cite{mu2024embodiedgpt}, the Embodied-Former module aligns embodied, visual, and textual information, effectively considering the agent's state and environmental information during task planning.
Unlike EmbodiedGPT, which directly uses third-person perspective images, LEO~\cite{huang2024embodied} encodes 2D egocentric images and 3D scenes into visual tokens. This method effectively perceives 3D world information and executes tasks accordingly.
Similarly, the EIF-Unknow model utilizes Semantic Feature Maps extracted from Voxel Features as visual tokens, which are input along with text tokens into a trained LLaVA model for task planning~\cite{Wu2024EmbodiedIF}.
Furthermore, embodied multimodal foundation models, or VLA models, have been extensively trained with large datasets in studies like the RT series~\cite{brohan2022rt, zitkovich2023rt}, PaLM-E~\cite{driess2023palm}, and Matcha~\cite{zhaoChatEnvironmentInteractive2023} to achieve alignment of visual and textual features in embodied scenarios. 

However, task planning is only the first step for an agent in completing an instruction task. Subsequent action planning determines whether the task can be accomplished. In the experiments from RoboGPT~\cite{chen2023robogpt}, the accuracy of task planning reached 96\%, but the overall task completion rate was only 60\%, limited by the performance of the low-level planner. Therefore, whether an embodied agent can transition from the cyber space of ``imagining how tasks are completed" to the physical world of ``interacting with the environment and completing tasks" hinges on effective action planning.

\subsection{Embodied Action Planning}
\label{sec:llp}
The distinction between task planning and action planning highlights that action planning must address real-world uncertainties due to the insufficient granularity of task planning subtasks~\cite{zhao2024survey}. Action planning can be achieved by: 1) using pre-trained embodied models to complete subtasks via APIs, or 2) leveraging the VLA model's capabilities. The results from action planning are fed back to refine task planning.

\subsubsection{Action utilizing APIs}
A common approach involves providing LLMs with definitions of well-trained policy models to understand and use them effectively for specific tasks~\cite{brohan2023can, Song2022LLMPlannerFG}. By generating code, LLMs can abstract tools into a function library, allowing better handling of sub-tasks~\cite{Liang2022CodeAP}. Reflexion adjusts these tools during execution to improve generalization~\cite{Shinn2023Reflexion}. DEPS enables LLMs to learn and combine various skills through zero-shot learning~\cite{Wang2023Describe}.
The hierarchical planning paradigm simplifies development by separating high-level task planning from specific action execution through policy models. This modularity allows independent development, testing, and optimization, enhancing flexibility and maintainability. While it enables adaptability to various tasks and environments, reliance on external policy models can introduce latency and affect performance, making the quality of these models crucial for overall agent effectiveness.

\subsubsection{Action utilizing VLA model}

Different form previous approach that task planning and action execution are performed within the same system, this paradigm leverages the capabilities of embodied multimodal foundation models for planning and executing actions, reducing communication latency and improving system response speed and efficiency. In VLA models, the tight integration of perception, decision-making, and execution modules allows the system to handle complex tasks and adapt to changes in dynamic environments more efficiently. This integration also facilitates real-time feedback, enabling the agent to self-adjust strategies, thereby enhancing the robustness and adaptability of task execution \cite{brohan2023rt, mu2024embodiedgpt, vuong2023open}.
However, this paradigm is undoubtedly more complex and costly, particularly when dealing with intricate or long-term tasks. Additionally, a key issue is that an action planner, without an embodied world model, cannot simulate physical laws using only the internal knowledge of an LLM. This limitation hinders the agent to accurately and effectively complete various tasks in the physical world, preventing the seamless transfer from cyber space to physical world.

\subsubsection{Scalability in Diverse Environments}

\red{Scalability in embodied agents involves adapting to increased complexity in larger and more diverse environments through robust perception, efficient decision-making, and resource optimization. Strategies include hierarchical SLAM for mapping, multi-modal perception, and energy-efficient edge computing. Collaborative scalability is enhanced via multi-agent systems and decentralized communication, while generalization relies on domain adaptation to operate in new environments. }

\section{Sim-to-Real Adaptation}





Sim-to-Real adaptation in embodied AI refers to the process of transferring capabilities or behaviors learned in simulated environments (cyber space) to real-world scenarios (physical world). It involves validating and improving the effectiveness of algorithms, models, and control strategies developed in simulation to ensure they perform robustly and reliably in physical environments. To achieve sim-to-real adaptation, embodied world models, data collection and training methods, and embodied control algorithms are three essential components.


\vspace{-5pt}
\subsection{Embodied World Model} 
\label{sub:EWM}



Sim-to-Real involves creating simulation-based world models that closely resemble real-world environments. These models predict the next state to make decisions and are trained from scratch on physical world data, unlike VLA models which are pre-trained on large-scale datasets and fine-tuned with real-world data. World models are effective for structured tasks like autonomous driving and object sorting but are less suited for unstructured, complex tasks.

\begin{figure}[!t]
    \centering
    \includegraphics[width=1\linewidth]{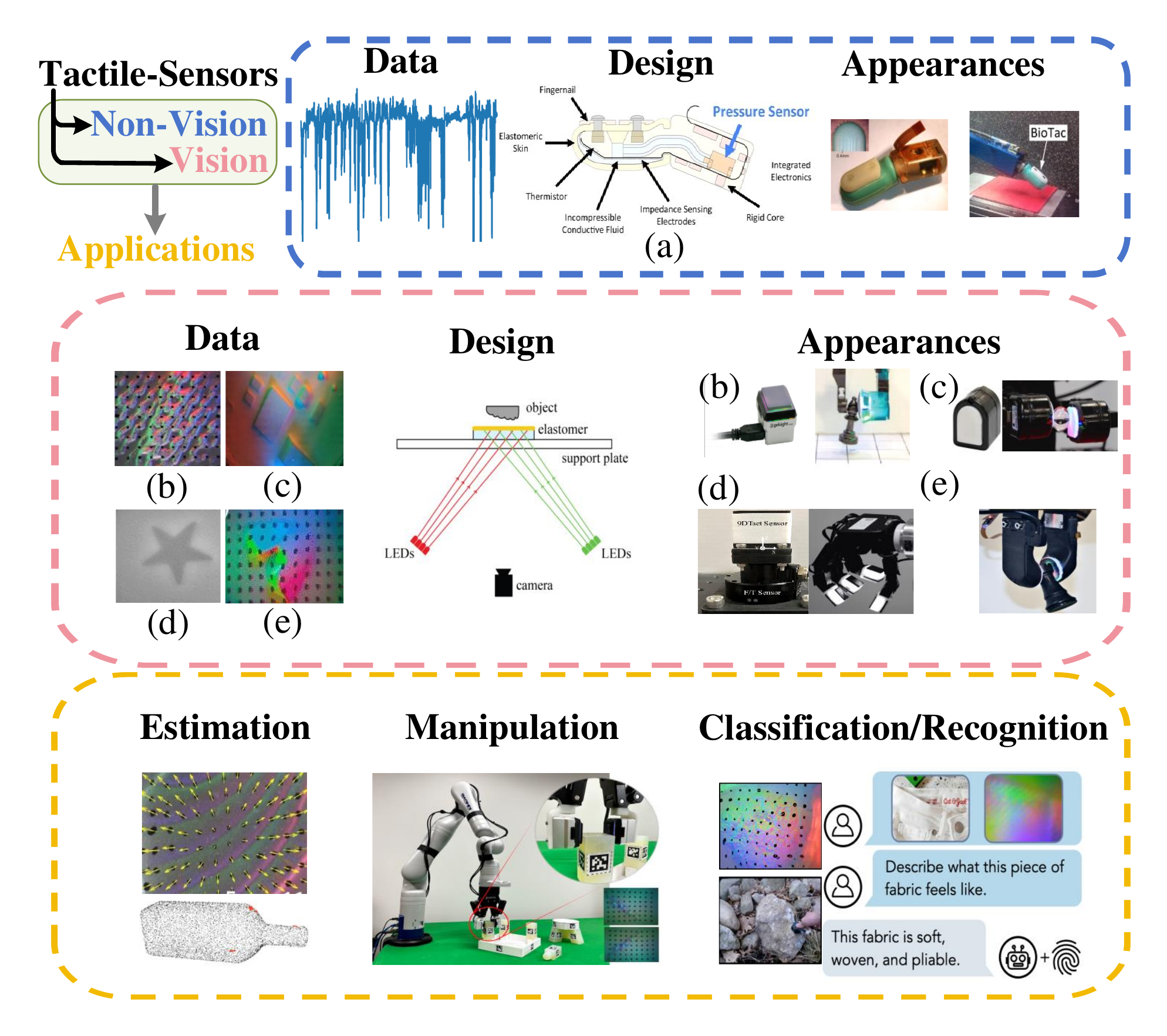}
    \vspace{-25pt}
    \caption{Embodied world models can be roughly divided into three type. (a) \textbf{Generation-based Methods} learn the transformation relation between the input space and the output space using an autoencoder framework. (b) \textbf{Prediction-based Methods} are more general frameworks where a world model is trained in latent space. (c) \textbf{Knowledge-driven Methods} inject artificially constructed knowledge into the model, giving the model world knowledge to obtain output that meets the given knowledge constraints. Note that the components within the dashed line are optional.}
        \vspace{-15pt}
    \label{fig:world-model-types}
\end{figure}

Learning world models is promising of the physical simulation field. Compared to traditional simulation methods, it offers significant advantages, such as the ability to reason about interactions with incomplete information, meet real-time computation requirements, and improve prediction accuracy over time. The predictive capability of such world models is crucial, enabling robots to develop the physical intuition necessary to operate in the human world. As shown in Fig. \ref{fig:world-model-types}, according to the learning pipeline of the world environment, they can be divided into Generation-based methods, Prediction-based methods and Knowledge-driven methods.


\subsubsection{Generation-based Methods}
As the scale of models and data increases, generative models have demonstrated the ability to understand and generate images (e.g., World Models \cite{ha2018world}), videos (e.g., Sora \cite{zhu2024sora}, Pandora \cite{xiang2024pandora}), point clouds (e.g., 3D-VLA \cite{zhen20243d}) or other formats of data (e.g., DWM \cite{ding2024diffusion}) that conform to physical laws. This capability suggests that generative models can internalize world knowledge. Specifically, after exposure to large datasets, these models not only capture statistical properties but also simulate physical and causal relationships through their intrinsic structures. Thus, generative models function as more than just pattern recognition tools; they exhibit characteristics of world models. The embedded world knowledge in these models can be harnessed to enhance the performance of other models. By leveraging this knowledge, we can improve model generalization, robustness, adaptability to new environments, and predictive accuracy on unseen data \cite{zhen20243d, xiang2024pandora}.
However, generative models also have certain limitations and drawbacks. For instance, they may produce inaccurate or distorted outputs when faced with biased data distributions or insufficient training data. Moreover, these models require substantial computational resources and time for training and often lack interpretability, which hinders practical application. While generative models have shown promise in generating content that adheres to physical laws, challenges such as improving efficiency, enhancing interpretability, and mitigating data bias must be addressed for broader application. Continued research is likely to unlock further value and potential in these models.


\subsubsection{Prediction-based Methods}




The prediction-based world model predicts and understands the environment by constructing and utilizing internal representations. By reconstructing corresponding features in the latent space based on provided conditions, it captures deeper semantics and associated world knowledge. This model maps input information to a latent space and operates within that space to extract and utilize high-level semantic information, thereby enabling the robots to perceive the essential representation of the world environment (e.g., I-JEPA\cite{assran2023self}, MC-JEPA\cite{bardes2023mc}, A-JEPA\cite{fei2023jepa}, Point-JEPA \cite{saito2024point}, IWM\cite{garrido2024learning}) and more accurately perform embodied downstream tasks (e.g., iVideoGPT\cite{wu2024ivideogpt}, IRASim\cite{zhu2024irasim}), STP\cite{yang2024spatiotemporal}, MuDreamer\cite{burchi2024mudreamer}). Latent features, unlike pixel-level information, can abstract and decouple various forms of knowledge, enabling models to handle complex tasks and scenes more effectively while enhancing generalization \cite{lecun2022path}. In spatiotemporal modeling, for instance, a world model predicts an object's post-interaction state by integrating its current state, the nature of the interaction, and internal knowledge. Specifically, embodied world models generate dynamic environmental predictions by combining perceptual information with prior knowledge, relying on both sensory data and inherent world knowledge to accurately infer and predict environmental changes \cite{wu2024ivideogpt, yang2024spatiotemporal, burchi2024mudreamer}. This process considers the current state of objects alongside their historical and contextual information.

Similarly, leveraging the world knowledge embedded in its representations can further enhance the model's perception and robustness\cite{assran2023self, bardes2023mc, dawid2023introduction, garrido2024learning}. By operating in latent space, it is expected that robots can maintain high performance in different environments at a lower cost\cite{burchi2024mudreamer}. The key to this approach lies in abstract processing and knowledge decoupling, enabling efficient adaptation to complex situations. However, such models may exhibit limitations and instability when dealing with previously unseen environments and conditions. Additionally, the world knowledge decoupled in the latent space may have interpretability issues.

\subsubsection{Knowledge-driven Methods}


\begin{figure*}[t]
    \centering
    \includegraphics[width=0.9\linewidth]{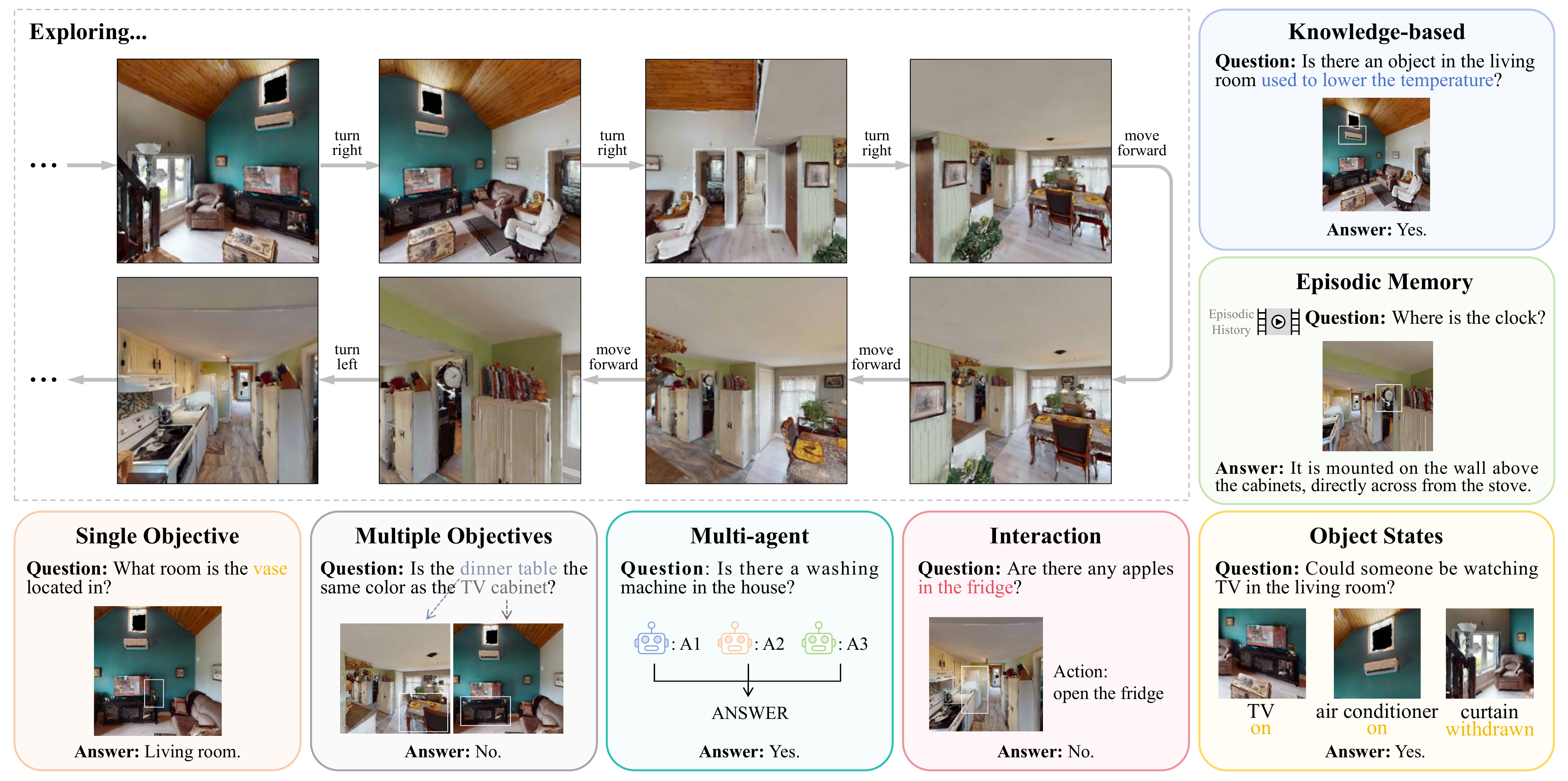}
    \vspace{-10pt}
    \caption{Five pipelines to achieve sim-to-real gap. \textbf{``Real2Sim2Real"} reduces the gap by reconstructing real scenes.  \textbf{``TRANSIC"} compensates for the sim-to-real transfer gap through human-corrected interventions. \textbf{``Domain Randomization"} enhances model transfer adaptability by simulating environmental diversity. \textbf{``System Identification"} improves sim-to-real environment similarity, thereby mitigating the sim-real gap. \textbf{``Lang4Sim2Real"} uses natural language to bridge two domains, learning invariant image representations and reducing visual gaps. }
        \vspace{-15pt}
    \label{fig:sim-to-real}
\end{figure*}

Knowledge-driven world models inject artificially constructed knowledge into the models, endowing them with world knowledge. This method has shown broad application potential in the field of embodied AI. For example, in the real2sim2real approach\cite{lim2022real2sim2real}, real-world knowledge is used to build physics-compliant simulators, which are then used to train robots, enhancing model robustness and generalization capabilities. Additionally, artificially constructing common sense or physics-compliant knowledge and applying them to generative models or simulators is a common strategy (e.g., ElastoGen \cite{feng2024elastogen}, One-2-3-45 \cite{liu2024one}, PLoT \cite{wong2023word}). This approach imposes more physically accurate constraints on the model, enhancing its reliability and interpretability in generation tasks. These constraints ensure the model's knowledge is both accurate and consistent, reducing uncertainty during training and application. Some approaches combine artificially created physical rules with LLMs or MLMs. By leveraging the commonsense capabilities of LLMs and MLMs, these approaches (e.g., Holodeck\cite{yang2024holodeck}, LEGENT\cite{cheng2024legent}, GRUtopia \cite{wang2024grutopia}) generate diverse and semantically rich scenes through automatic spatial layout optimization. This greatly advances the development of general-purpose embodied agents by training them in novel and diverse environments. 



\subsubsection{Limitations}\red{Current limitations of world models include handling the complexity and variability of real-world environments, such as high-dimensional sensory inputs, dynamic and stochastic elements, and long-term dependencies. Many models struggle with generalization across unseen scenarios, often requiring extensive training data and computational resources. Additionally, sim-to-real transfer remains problematic, as simulated environments fail to fully capture real-world physics and noise. These limitations can be addressed by integrating more realistic simulators, incorporating multimodal sensory inputs, and using hierarchical or modular architectures. Improving data efficiency, enhancing transfer learning techniques, and incorporating real-world priors can also enable more accurate predictions and adaptive decision-making. Recently, the Cosmos Platform \cite{agarwal2025cosmos} was proposed by NVIDIA, which contains autoregressive and diffusion models for Text-to-World and Video-to-World generation, to accelerate the development of physical AI systems. It may give us some inspirations about how to build an embodied world model.
}

\subsection{Data Collection and Training}
For sim-to-real adaptation, the high-quality data is important. Traditional data collection methods involve expensive equipment and precise operations, which are time-consuming and labor-intensive. Recently, some efficient and cost-effective methods have been proposed for high-quality demonstration data collection and training. We discuss data collection methods in both real-world and simulated environments. 


\subsubsection{Real-World Data}
Training large, high-capacity models with rich datasets has shown great success. This approach is also promising for robotics, where large, diverse datasets can enhance generalization and adaptability. Open X-Embodiment \cite{vuong2023open} provides data from 22 robots with 527 skills and 160,266 tasks in domestic settings. UMI \cite{chi2024universal} offers a framework for collecting dynamic, bimanual data using a handheld gripper. Mobile ALOHA \cite{fu2024mobile} enables data collection for full-body mobile manipulation tasks. Human-agent collaboration \cite{luo2024human} improves data quality and efficiency by combining human input with agent refinement processes.
\subsubsection{Simulated Data}
Data collection in real-world settings is resource-intensive and time-consuming, making simulation-based data collection an attractive alternative. Simulated environments allow for automated, efficient data collection. For example, CLIPORT \cite{shridhar2022cliport} and Transporter Networks \cite{zeng2021transporter} utilized Pybullet simulator data for training and successfully transferred models to real-world applications. GAPartNet \cite{geng2023gapartnet} developed a large-scale dataset with detailed part-level annotations for better object interaction in both simulations and reality. SemGrasp \cite{li2024semgrasp} created the CapGrasp dataset for semantically rich hand grasping in virtual environments.

\subsubsection{Sim-to-Real Paradigms}


Recently, several sim-to-real paradigms have been introduced, to mitigate the need for extensive and costly real-world demonstration data by conducting extensive learning in simulation environments, followed by migration to real-world settings. This section outlines five paradigms for sim-to-real transfer, as shown in Fig. \ref{fig:sim-to-real}.

Real2Sim2real \cite{torne2024reconciling} improves imitation learning by using reinforcement learning in a "digital twin" simulation to develop strategies, which are then transferred to the real world. TRANSIC \cite{jiang2024transic} reduces the sim-to-real gap through real-time human intervention and residual policy training based on corrected behaviors. Domain Randomization \cite{tobin2017domain, andrychowicz2020learning, matas2018sim} increases model generalization by varying simulation parameters to cover real-world conditions. System Identification \cite{kaspar2020sim2real, yu2017preparing} creates accurate simulations of real-world scenes to ensure smooth transitions from simulation to reality. Lang4sim2real \cite{yu2024natural} leverages natural language descriptions to bridge the sim-to-real gap, improving model generalization with cross-domain image representations.

\begin{figure}[!t]
    \centering
    \includegraphics[width=0.9\linewidth]{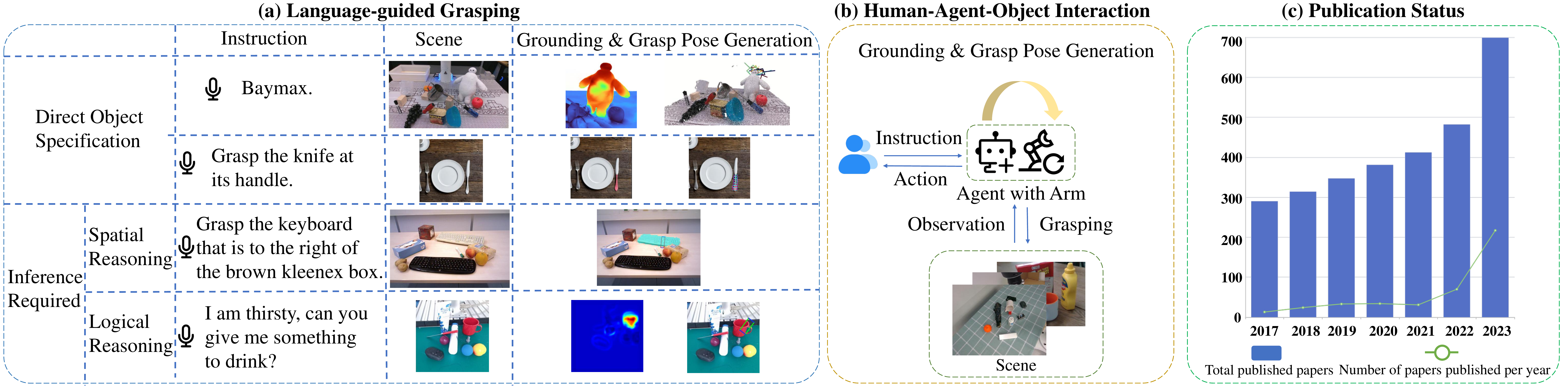}
        \vspace{-10pt}
    \caption{Exemplar tasks from ARIO, where the top row indicates the task category while the text at the bottom row provides task instructions.}
        \vspace{-12pt}
\label{fig:ario}
\end{figure}

\subsubsection{ARIO (All Robots in One)}

Despite the seemingly unified structure of pre-training datasets like Open X-embodiment, several critical issues remain unresolved. These issues include the absence of comprehensive sensory modalities—no dataset currently integrates images, 3D vision, text, tactile, and auditory inputs simultaneously. Additionally, the lack of a unified format in multi-robot datasets complicates data processing and loading. Furthermore, there is an incompatibility in representing diverse control objects across different robotic platforms, insufficient data volume that hinders large-scale pretraining, and a scarcity of datasets that combine both simulated and real data, which is essential for addressing the sim-to-real gap.

To overcome these challenges, we propose ARIO (All Robots In One) , which is a new dataset standard that optimizes existing datasets and facilitates the development of more versatile and general-purpose embodied agents. The ARIO standard \cite{wang2024all} records control and motion data from robots with different morphologies in a unified format. ARIO's unified format accommodates variable data from diverse robot types, ensuring precise timestamps. This standard enables users to efficiently train high-performing, generalizable embodied AI models, positioning ARIO as the ideal format for embodied AI datasets. Building on the ARIO standard, a unified large-scale ARIO dataset is further developed,  which comprises approximately 3 million episodes collected from 258 series and 321,064 tasks. Fig. \ref{fig:ario} shows exemplar tasks from ARIO.

The ARIO dataset addresses the limitations of current datasets and facilitates the development of robust, general-purpose embodied agents. By providing a cohesive framework for data collection and representation, ARIO paves the way for the development of more powerful and versatile embodied agents, capable of navigating and interacting with the physical world in complex and diverse ways.


\subsubsection{Real-world Deployments of Embodied AI Systems}
\red{Embodied AI systems have made significant strides across various scenes. In healthcare, robots like the Da Vinci Surgical System and Moxi automate tasks such as surgery precision and supply deliveries. In logistics, Amazon Robotics and Boston Dynamics’ Stretch improve efficiency in warehousing and transportation. Manufacturing benefits from AI-driven robots like Fanuc and ABB, enhancing precision and collaboration. Nevertheless, sim-to-real adaptation faces significant challenges, including the domain gap between simulation and real-world data distributions, the complexity of dynamic real-world interactions, and the limited diversity of training data. Models often overfit to simulations, struggle with real-world sensor noise, and fail to handle unexpected events.}

\section{Challenges and Future Directions}

Despite of the rapid progress of embodied AI, it faces several challenges and presents exciting future directions.

\red{\textbf{High-quality Robotic Datasets}: Obtaining sufficient real world robotic data remains a significant challenge. Collecting this data is both time-consuming and resource-intensive. Relying solely on simulation data worsens the sim-to-real gap problem. Creating diverse real world robotic datasets necessitates close and extensive collaboration among various institutions. Additionally, the development of more realistic and efficient simulators is essential for improving the quality of simulated data.  For building generalizable embodied models capable of cross-scenario and cross-task applications in robotics, it is essential to construct large-scale datasets, leveraging high-quality simulated environment data to assist real world data.}

\red{\textbf{Long-Horizon Task Execution}: Executing single instructions can often entail long-horizon tasks for robots, exemplified by commands like ``clean the kitchen", which involve activities such as rearranging objects, sweeping floors, wiping tables, and more. Accomplishing such tasks successfully necessitates the robot's ability to plan and execute a sequence of low-level actions over extended time spans. While current high-level task planners have shown initial success, they often prove inadequate in diverse scenarios due to their lack of tuning for embodied tasks. Addressing this challenge requires the development of efficient planners equipped with robust perception capabilities and much commonsense knowledge. To balance the trade-off between planning complexity and real-time adaptability, we can combine lightweight monitor modules for high-frequency monitoring, and two adapters for subtask and path adaptation reasoning at a lower frequency.  }

\red{\textbf{Causal Reasoning}: Existing data-driven embodied agents make decisions based on intrinsic data correlations. However, this approach does not allow the models to truly understand the causal relations between knowledge, behavior, and environment, resulting in biased strategies. This makes it difficult to ensure that they can operate in real-world environments in a robust and reliable manner. Therefore, it is important for embodied agents to be driven by world knowledge, capable of autonomous causal reasoning. By understanding the world through interaction and learning its workings via abductive reasoning, we can further enhance the adaptability, decision reliability, and generalization capabilities of  embodied agents in complex real-world environments. For embodied tasks, it is necessary to establish spatial-temporal causal relations across modalities through interactive instructions and state predictions. Moreover, agents need to understand the affordances of objects to achieve adaptive task planning in dynamic scenes. }



\red{\textbf{Unified Evaluation Benchmark}: While numerous benchmarks exist for evaluating low-level control policies, they vary significantly in the assessed skills. Furthermore, the objects and scenes in these benchmarks are typically limited by simulator constraints. To comprehensively evaluate embodied models, the benchmark should encompass a diverse range of skills using realistic simulators. Many benchmarks for high-level task planners focus on assessing planning capability through question-answering tasks. However, a more desirable approach involves evaluating both the high-level task planner and the low-level control policy together for executing long-horizon tasks and measuring success rates, rather than relying solely on isolated assessments of the planner.}

\red{\textbf{Security and Privacy:}  Embodied agents face significant security challenges when deployed in sensitive or private spaces. These agents often rely on LLMs for decision-making, which introduces new vulnerabilities. For instance, LLMs are susceptible to backdoor attacks like word injection, scenario manipulation, and knowledge injection, which can lead to dangerous outcomes such as autonomous vehicles accelerating towards obstacles or robots performing hazardous actions. To mitigate these risks, we can evaluate potential attack vectors and develop more robust defenses. Additionally, the secure prompting, state management, and safety validation mechanisms can be used to enhance the security and robustness.}

\section{Conclusion}

\red{Embodied AI allows agents to sense, perceive, and interact with various objects from both cyber space and physical world, which exhibits its vital significance toward achieving AGI. This survey extensively reviews embodied robots, simulators, four representative embodied tasks: visual active perception, embodied interaction, embodied agents and sim-to-real adaptation, and future research directions. The comparative summary of the embodied robots, simulators, datasets, and approaches provides a clear picture of the recent development in embodied AI, which greatly benefits the future research along this emerging and promising research direction.}


\bibliographystyle{IEEEtran}
\bibliography{IEEEabrv,bibfile}

\begin{IEEEbiography}[{\includegraphics[width=1in,height=1.25in,clip,keepaspectratio]{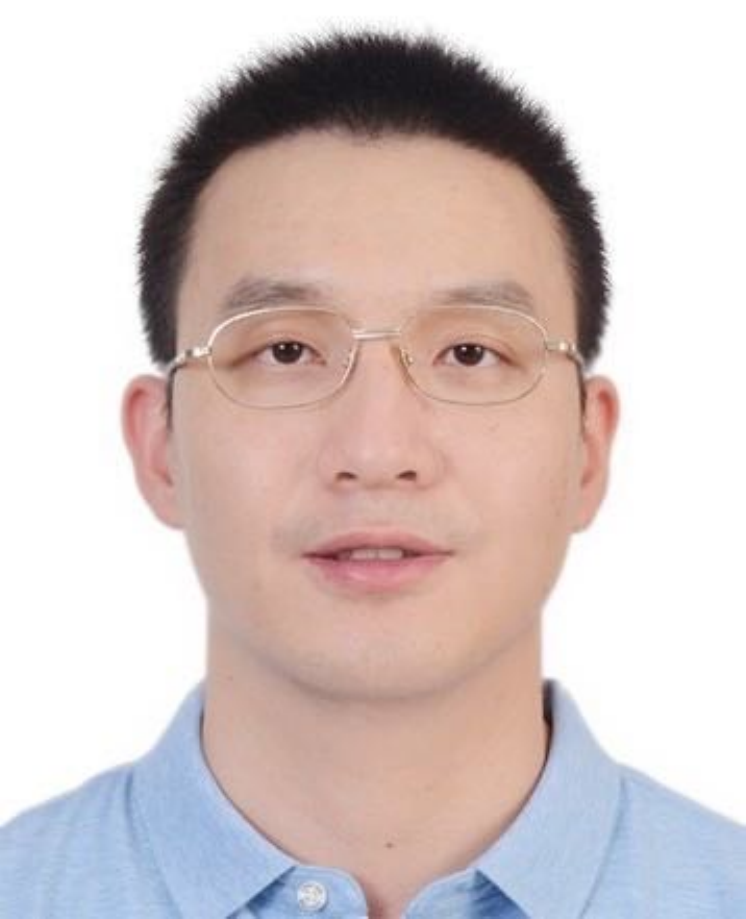}}]{Yang Liu}(M'21) is currently an Associate Professor working at the School of Computer Science and Engineering, Sun Yat-sen University. He received his Ph.D. degree from Xidian University in 2019. His current research interests include multi-modal reasoning, causality learning and embodied AI. He is the recipient of the First Prize of the Third Guangdong Province Young Computer Science Academic Show. He has authorized and co-authorized more than 40 papers in top-tier academic journals and conferences such as TPAMI, TIP, CVPR and ICCV.
\end{IEEEbiography}

\begin{IEEEbiography}[{\includegraphics[width=1in,height=1.25in,clip,keepaspectratio]{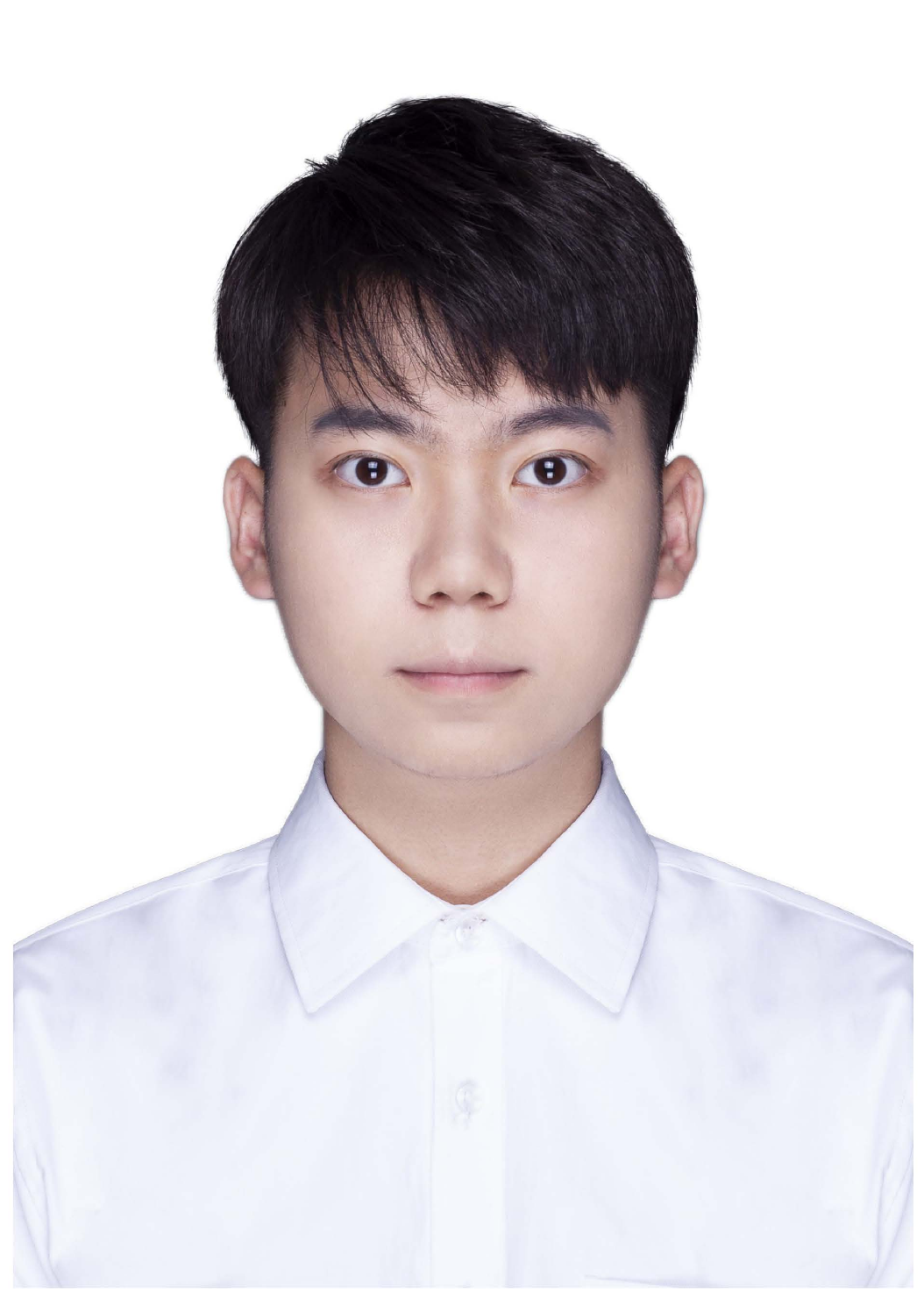}}]{Weixing Chen} has received the B.S. degree from the college of Medicine and Biological Information Engineering, Northeastern University, in 2020 and M.S. degree from Shenzhen Institute of Advanced Technology, Chinese Academy of Sciences in 2023. He is currently a Ph.D. student at the School of Computer Science and Engineering, Sun Yat-sen University. His main interests include multi-modal learning, causal relation discovery, and embodied ai. He has been serving as a reviewer for numerous academic journals and conferences such as TNNLS, NeurIPS, ICML, ICLR, MICCAI, and ACM MM.
\end{IEEEbiography}

\begin{IEEEbiography}[{\includegraphics[width=1in,height=1.25in,clip,keepaspectratio]{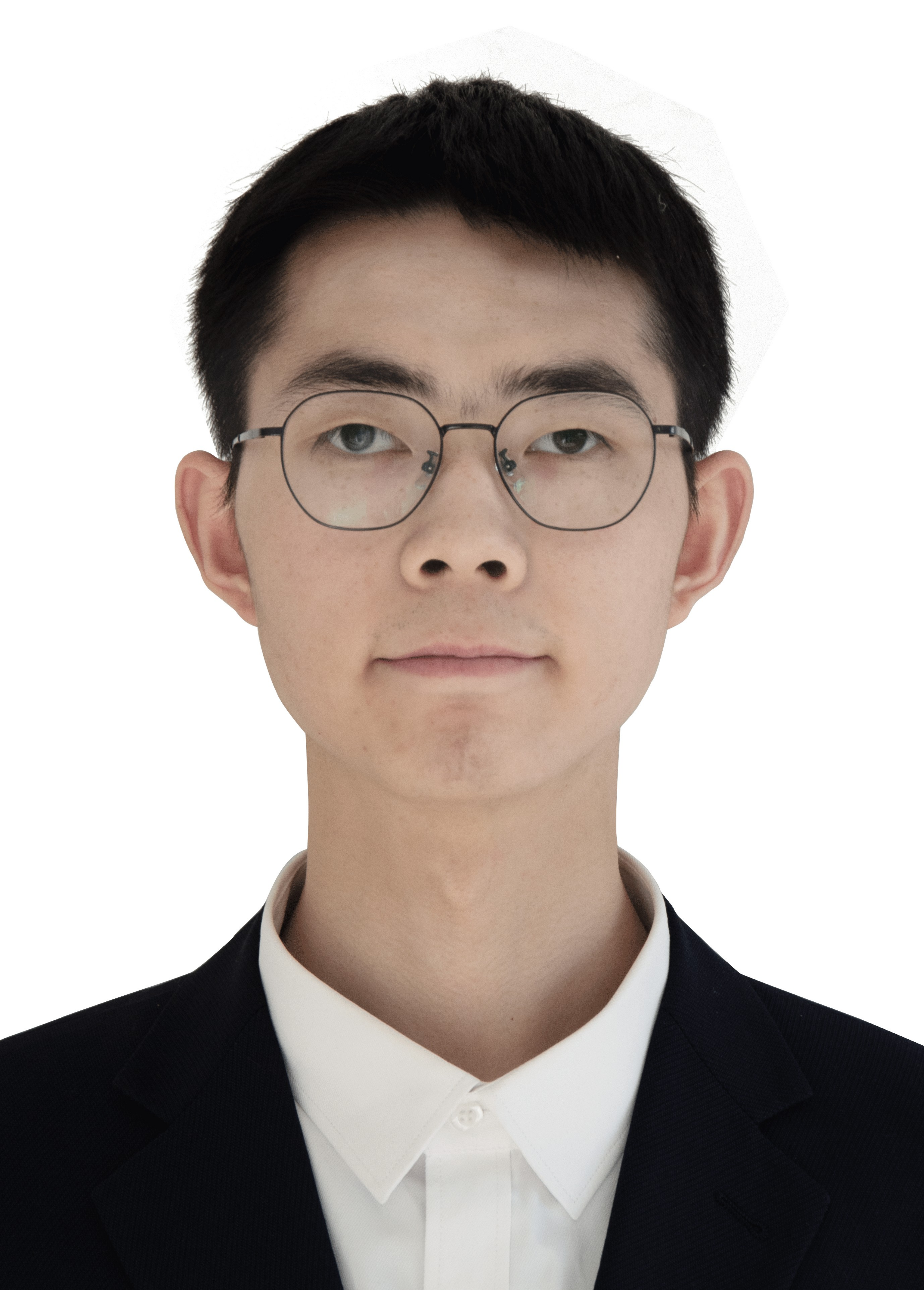}}]{Yongjie Bai} received the B.S. degree from the School of Computer Science and Technology, Dalian University of Technology, in 2024. He is currently a Ph.D. student at the School of Computer Science and Engineering, Sun Yat-sen University. His main research interests include embodied AI, robot learning, and multi-modal learning.
\end{IEEEbiography}

\begin{IEEEbiography}[{\includegraphics[width=1in,height=1.25in,clip,keepaspectratio]{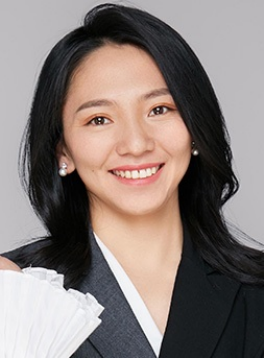}}]{Xiaodan Liang}(Senior Member, IEEE) is currently a Professor at Sun Yat-sen University. She was a postdoc researcher in the machine learning department at Carnegie Mellon University, working with Prof. Eric Xing, from 2016 to 2018. She received her PhD degree from Sun Yat-sen University in
2016, advised by Liang Lin. She has published several cutting-edge projects on human-related analysis, including human parsing, pedestrian detection and instance segmentation, 2D/3D human pose estimation, and activity recognition.
\end{IEEEbiography}

\begin{IEEEbiography}[{\includegraphics[width=1in,height=1.25in,clip,keepaspectratio]{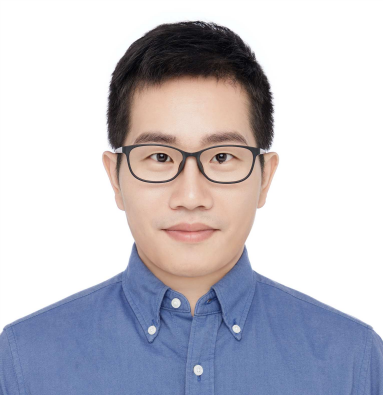}}]{Guanbin Li}(M'15) is currently a Professor in School of Computer Science and Engineering, Sun Yat-Sen University. He received his PhD degree from the University of Hong Kong in 2016. His current research interests include computer vision, image processing, and deep learning. He is a recipient of ICCV 2019 Best Paper Nomination Award. He has authorized and co-authorized on more than 100 papers in top-tier academic journals and conferences. He serves as an area chair for the conference of VISAPP. He has been serving as a reviewer for numerous academic journals and conferences such as TPAMI, IJCV, TIP, TMM, TCyb, CVPR, ICCV, ECCV and NeurIPS.
\end{IEEEbiography}

\begin{IEEEbiography}[{\includegraphics[width=1in,height=1.25in,clip,keepaspectratio]{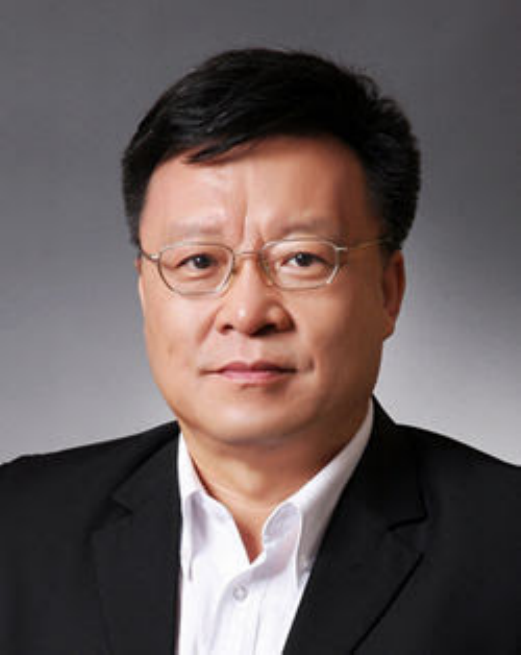}}]{Wen Gao}(Fellow, IEEE) received the Ph.D. degree in electronics engineering from the
University of Tokyo, Tokyo, Japan, in 1991. He is currently a Professor of computer science with the School of Electronic Engineering and Computer Science, Institute of Digital Media, Peking
University, Beijing, China. Before joining Peking University, he was a Professor of computer science with the Harbin Institute of Technology, Harbin, China, from 1991 to 1995, and a Professor with the Institute of Computing Technology, Chinese Academy of Sciences, Beijing, China. He has authored or coauthored extensively, including five books and more than 600 technical articles in refereed journals and conference proceedings in the areas of image processing, video coding and communication, pattern recognition, multimedia information retrieval, multimodal interfaces, and bioinformatics. He is a Member of the China Engineering Academy. 
\end{IEEEbiography}

\begin{IEEEbiography}[{\includegraphics[width=1in,height=1.25in,clip,keepaspectratio]{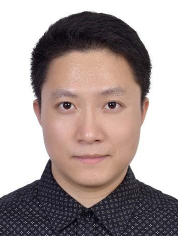}}]{Liang Lin}(Fellow, IEEE) is currently a Full Professor with Sun Yat-sen University, Guangzhou, China. From 2008 to 2010, he was a Postdoctoral Fellow with the University of California, Los Angeles, Los Angeles, CA, USA. From 2016 to 2018, he led the SenseTime R\&D teams to develop cutting-edge and deliverable solutions for computer vision, data analysis and mining, and intelligent robotic systems. He has authored or coauthored more than 100 papers in top-tier academic journals and conferences, such as 15 papers in IEEE TRANSACTIONS ON PATTERN ANALYSIS AND MACHINE I NTELLIGENCE and International Journal of Computer Vision, and more than 60 papers in CVPR, ICCV, NeurIPS, and IJCAI. He was an Associate Editor for IEEE TRANSACTIONS ON MULTIMEDIA , IEEE TRANSACTIONS ON NEURAL NETWORKS AND LEARNING SYSTEMS, and was an Area/Session Chair for numerous conferences, such as CVPR, ICCV, AAAI, ICME, and ICMR. He was the recipient of the Annual Best Paper Award by Pattern Recognition (Elsevier) in 2018, Best Paper Diamond Award at IEEE ICME 2017, Best Paper Runner-Up Award at ACM NPAR 2010, Google Faculty Award in 2012, Best Student Paper Award at IEEE ICME 2014, and Hong Kong Scholars Award in 2014. He is a Fellow of IAPR, AAIA, and IET.
\end{IEEEbiography}

\end{document}